\pdfoutput=1

\documentclass[11pt]{article}
\usepackage[figuresright]{rotating}
\usepackage[]{acl}
\usepackage{amsmath}
\usepackage{times}
\usepackage{latexsym}
\usepackage{subcaption}
\usepackage{tabularx}
\usepackage{latexsym}
\usepackage{multirow}
\usepackage{algorithm}
\usepackage{algpseudocode}
\usepackage{array}

\usepackage[T1]{fontenc}

\usepackage{ulem}
\usepackage{soul}
\usepackage{enumitem}
\usepackage[utf8]{inputenc}
\usepackage{xspace}
\usepackage{booktabs}

\usepackage{microtype}
\definecolor{myblue}{RGB}{6, 82, 221}
\definecolor{myorange}{RGB}{211, 84, 0}
\definecolor{lowblue}{RGB}{102,178,255}
\definecolor{justblue}{RGB}{84, 160, 255}
\definecolor{mypurple}{RGB}{108, 92, 231}
\definecolor{mygray}{RGB}{158, 158, 158}
\definecolor{lowpurple}{RGB}{204,153,255}
\definecolor{lowwhite}{RGB}{255,255,255}
\definecolor{verylowpurple}{RGB}{255,102,102}
\definecolor{embcolor}{RGB}{255,255,255}
\definecolor{myred}{RGB}{235, 47, 6} 
\definecolor{mygreen}{RGB}{162, 217, 206} 
\definecolor{fontgrey}{RGB}{44, 62, 80}
\definecolor{lowpurple}{RGB}{210, 180, 222}
\definecolor{mypumpkin}{RGB}{229, 152, 102}
\definecolor{lowgreen}{RGB}{171, 235, 198}
\definecolor{lowgreen2}{RGB}{186, 220, 88}
\definecolor{lowred}{RGB}{245, 183, 177}
\definecolor{lowyellow}{RGB}{241, 196, 15}
\definecolor{mypink}{RGB}{255, 118, 117}
\definecolor{bluemartina}{RGB}{18, 203, 196}
\definecolor{puffin}{RGB}{250, 152, 58}
\definecolor{grass}{RGB}{0, 148, 50}
\definecolor{cnngray}{RGB}{116, 125, 140}
\usepackage{color, colortbl}
\definecolor{Gray}{gray}{0.9}
\usepackage{graphicx}

\newcommand{\timeqa}{\textsc{Time\-QA}\xspace}
\newcommand{\templama}{\textsc{Temp\-LAMA}\xspace}
\newcommand{\tempreason}{\textsc{Temp\-Reason}\xspace}

\newcommand{\cotempqa}{\textsc{CoTemp\-QA}\xspace}

\definecolor{color1}{rgb}{0.1,0.7,0.8}
\definecolor{color2}{rgb}{0.9,0.1,0.1}
\definecolor{color3}{rgb}{0.7,0.3,0.7}
\definecolor{color4}{rgb}{0.3,0.3,0.7}
\definecolor{color5}{RGB}{8, 102, 3}
\definecolor{color6}{rgb}{0.53, 0.66, 0.42}

\title{Living in the Moment:\\ Can Large Language Models Grasp Co-Temporal Reasoning?}

\author{Zhaochen Su$^{1}$\thanks{Work was done during the internship at Shanghai AI lab.}, Juntao Li$^{1}$\thanks{Juntao Li is the Corresponding Author.}, Jun Zhang$^1$, Tong Zhu$^1$,\\ \textbf{Xiaoye Qu}$^{2,3}$, \textbf{Pan Zhou}$^3$, \textbf{Bowen Yan}$^4$, \textbf{Yu Cheng}$^5$, \textbf{Min Zhang}$^{1}$   \\  
 $^{1}$Institute of Computer Science and Technology, Soochow University, China \\
 $^{2}$Shanghai AI Laboratory, $^{3}$Huazhong University of Science and Technology\\
 $^{4}$Tsinghua University, $^{5}$The Chinese University of Hong Kong \\
 \texttt{\{suzhaochen0110,junzhang20030309\}@gmail.com}; \\
\texttt{\{ljt,minzhang\}@suda.edu.cn}; \texttt{tzhu1997@outlook.com} \\ \texttt{\{xiaoye,panzhou\}@hust.edu.cn}; \texttt{chengyu@cse.cuhk.edu.hk}  \\
 }

\begin{document}
\maketitle
\begin{abstract}
Temporal reasoning is fundamental for large language models~(LLMs) to comprehend the world.
Current temporal reasoning datasets are limited to questions about single or isolated events, falling short in mirroring the realistic temporal characteristics involving concurrent nature and intricate temporal interconnections.
In this paper, we introduce \textsc{CoTemp\-QA}, a comprehensive co-temporal Question Answering (QA) benchmark containing four co-temporal scenarios~(Equal, Overlap, During, Mix) with 4,748 samples for evaluating the co-temporal comprehension and reasoning abilities of LLMs.
Our extensive experiments reveal a significant gap between the performance of current LLMs and human-level reasoning on \textsc{CoTemp\-QA} tasks. Even when enhanced with Chain of Thought (CoT) methodologies, models consistently struggle with our task.
In our preliminary exploration, we discovered that mathematical reasoning plays a significant role in handling co-temporal events and proposed a strategy to boost LLMs' co-temporal reasoning from a mathematical perspective.
We hope that our \textsc{CoTemp\-QA} datasets will encourage further advancements in improving the co-temporal reasoning capabilities of LLMs.
Our code is available at \url{https://github.com/zhaochen0110/Cotempqa}.
\end{abstract}

\section{Introduction}
Recent advanced Large Language Models (LLMs) like GPT-4~\cite{openai2023gpt4} have shown impressive capabilities in understanding, generating, and reasoning about natural language~\cite{wei2022finetuned, zhao2023survey, chang2023survey}.
Despite their advancements, these models fall short in mastering temporal reasoning~\cite{chu2023timebench}, which is fundamental for humans to comprehend the world and distinguish daily events~\cite{chen2021a, su-etal-2023-efficient, tan-etal-2023-towards}, requiring a complex integration of capabilities, involving implicit arithmetic calculations~\cite{zhu-etal-2023-solving}, understanding logical implications~\cite{wei2022chain}, and leveraging extensive world knowledge~\cite{chu2023timebench}.

\begin{figure}[tb]
    \centering
    \includegraphics[width=1\linewidth]{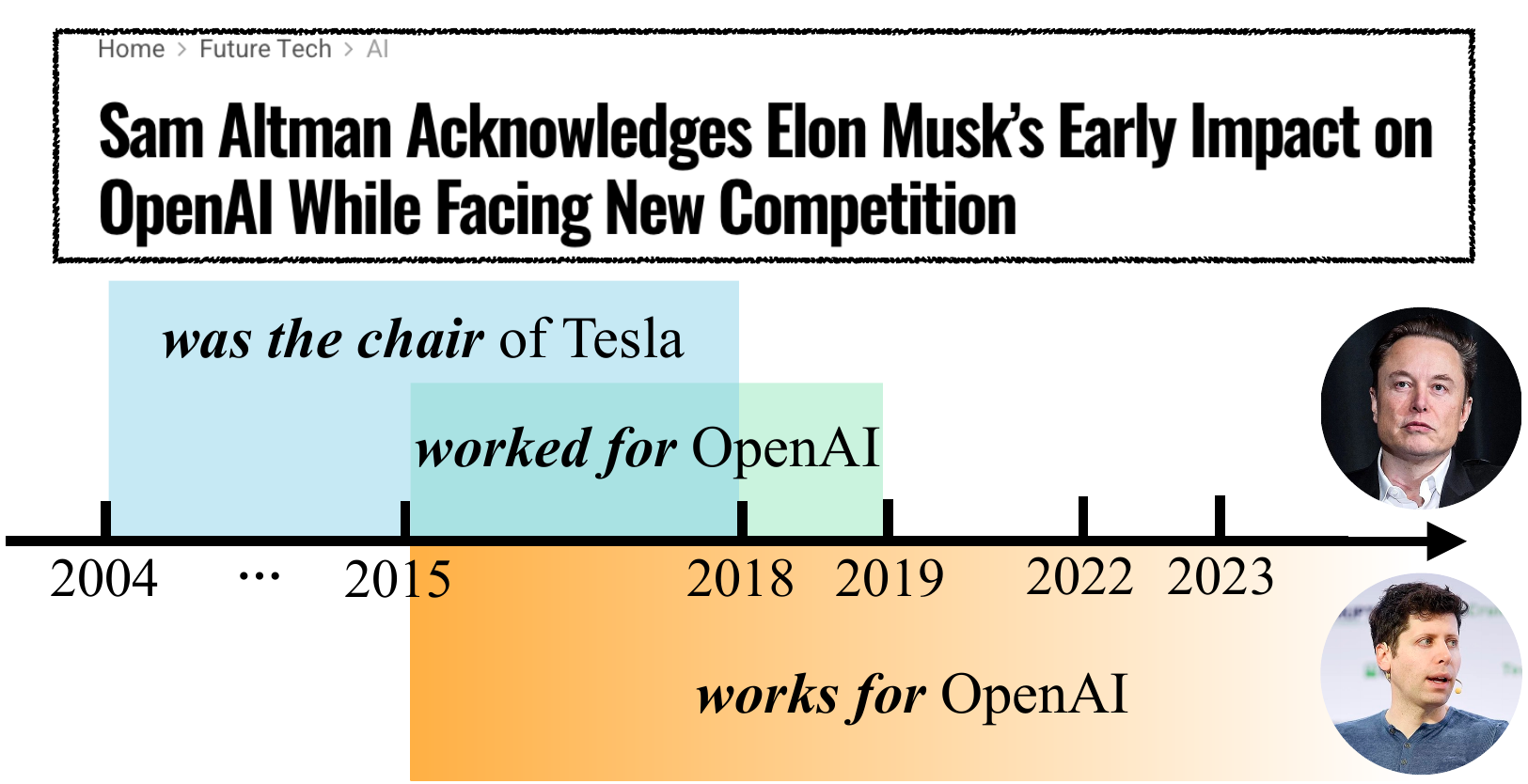}
    \caption{
    Understanding concurrent is crucial for us to understand how individuals navigate and influence diverse aspects of real-world scenarios. For instance, when \textit{Elon Musk} was the chair of \textit{Tesla}, he also worked for \textit{OpenAI}. Concurrently, \textit{Sam Altman} was working for \textit{OpenAI}, too. Their simultaneous experiences greatly influenced subsequent decision-making at \textit{OpenAI}.}
\label{biden's carrer}
\end{figure}

\begin{table*}[t]
\centering
\resizebox{\linewidth}{!}{
\begin{tabular}{llc}
\toprule
\textbf{Datasets} & \textbf{Question} & \textbf{Answer}  \\
\midrule

\multirow{2}{*}{\textbf{\timeqa}~\shortcite{chen2021a}}  &  Which school did Sam Altman attended in 2005? & Stanford University \\
& Which position did Elon Musk hold in 2005? & chairman of Tesla\\ 
\specialrule{0.04em}{1pt}{1pt}

\multirow{2}{*}{\textbf{\templama}~\shortcite{dhingra-etal-2022-time}}   &  In 2005, Sam Altman attended \_X\_. & Stanford University \\
& In 2005, Elon Musk hold the position of \_X\_.  & chairman of Tesla \\ 
\specialrule{0.04em}{1pt}{1pt}

\multirow{2}{*}{\textbf{\tempreason}~\shortcite{tan-etal-2023-towards}}   &  Which school did Sam Altman attend before he held the position of president of Y Combinator? & Standford University \\
& Which position did Elon Musk held before he worked for OpenAI?  & chairman of Tesla\\
\specialrule{0.04em}{1pt}{1pt}

\multirow{2}{*}{\textbf{\cotempqa~(ours)}}   &  When Elon Musk was working for OpenAI, where did he work for within the same time interval?~(\textbf{Overlap}) & Tesla, SpaceX \\
& While Elon Musk was working for OpenAI, where did Sam Altman work for concurrently?~(\textbf{During})  & OpenAI \\
 \bottomrule
\end{tabular}}
\caption{Example questions of prior TSQA datasets and our \cotempqa datasets.}
\label{tab:datasets}
\end{table*}

Current studies in temporal reasoning mainly focus on time-sensitive question-answering~(TSQA).
\citet{chen2021a} first introduced the \timeqa datasets, constructing time-evolving facts for a given subject and formulating questions based on the specific timestamp within the evolutionary facts.
\templama~\cite{dhingra-etal-2022-time} extracted structured facts from the Wikidata Knowledge Base~\cite{Wikidata} for closed-book TSQA.
Furthermore, \tempreason~\cite{tan-etal-2023-towards} translated explicit temporal expressions into the implicit event information within questions, offering a more comprehensive evaluation framework of TSQA.
Given the fact \textit{``Elon musk held the position of Tesla's chairman from 2004 to 2018''}, the models are tasked with accurately interpreting and responding to time specifiers in the questions, i.e., \textit{``Which position did Elon Musk hold in 2005?''} in \timeqa~\cite{chen2021a} or \textit{``Which position did Elon Musk held before he worked for OpenAI?''} in \tempreason~\cite{tan-etal-2023-towards}.

\begin{table}[t]
\centering
\small
\begin{tabular}{ll}
\hline
\textbf{Interpretation\qquad\qquad} & \textbf{Relation}\\[2pt] \hline
$\bullet \mbox{------} x \mbox{------} \bullet$   & \multirow{2}{*}{$x$ \textit{is equal to} $y$}  \\
$\circ \mbox{------} y \mbox{------} \circ$ \\[4pt] \hline

$\bullet \mbox{---} x \mbox{---} \bullet$  & \multirow{2}{*}{$x$ \textit{overlaps with} $y$} \\
$~~~~~~~~\circ \mbox{---} y \mbox{---} \circ$ \\ \hline

$~~~\bullet \mbox{---} x \mbox{---} \bullet$  & \multirow{2}{*}{$x$ \textit{during} $y$}  \\
$\circ \mbox{------} y \mbox{------} \circ$ \\ \hline

\end{tabular}
\caption{Interpretation of three co-temporal relations.}
\label{tab:allen}
\end{table}

The datasets mentioned above provide a straightforward way to evaluate LLMs' capabilities in temporal reasoning.
However, as LLMs evolve, there is an urgent need to evaluate their proficiency in more realistic scenarios.
As shown in Figure~\ref{biden's carrer}, the reality might present a more intricate and multifaceted nature, involving concurrent events and complex temporal interconnections over time~\cite{DBLP:journals/corr/abs-1206-5333}.
Current datasets mainly question single or isolated events and might not fully reflect the realistic temporal characteristics.
Therefore, we create the Co-Temporal QA~(\cotempqa) datasets to complement existing corpora by focusing on the concurrent nature of time and co-temporal relations in real-world situations.
Experiments conducted on both closed-book and open-book QA settings across 14 large language models reveal that even the advanced model GPT-4 is well below a satisfactory co-temporal reasoning performance.
Specifically, GPT-4 achieves an overall score of 54.7, and the best open-source LLM is 30.1, which significantly falls behind the human performance of 92.8.
We also observe that the representative reasoning enhancement strategies, e.g., Chain-of-Thought (CoT)~\cite{NEURIPS2022_9d560961}, fail to consistently improve and even reduce the temporal reasoning capabilities of LLMs in some scenarios.

Throughout the investigation on our \cotempqa, we observed that mathematical reasoning plays a crucial role in handling co-temporal events.
Building on this insight, we propose a simple but effective \textsc{Math-reasoning} CoT~(\textsc{Mr-CoT}) strategy to boost the co-temporal reasoning capability of LLMs, achieving a remarkable 10.8 point improvement over existing baselines.
However, it is important to note that there remains a nonnegligible gap between the performance of our proposed \textsc{Mr-CoT} and human-level reasoning in handling complex, concurrent temporal relations.
We hope our research could inspire more great works to improve the co-temporal ability of LLMs.

\begin{table}[t]
\resizebox{\columnwidth}{!}{
\begin{tabular}{lcccc}
\toprule
\textbf{Mode}                    & \textbf{Questions}    & \textbf{Subjects}     & \textbf{\#Facts}       & \textbf{\#Answers}      \\
\midrule
\textbf{Equal}            &436
                 & 401      & 11.65       & 1.17       \\
\textbf{Overlap}            & 653
       & 591      & 14.51      & 1.23      \\
\textbf{During}            & 3,096 & 2,161      & 15.05       & 1.33       \\
\textbf{Mix}            & 563 & 434      & 12.54      & 2.27       \\ \hline
\textbf{Total}            & 4,748 & 3,587      & 14.45      & 1.41       \\
\bottomrule
\end{tabular}}
\caption{Statistics of our datasets.  \#Facts and \#Answers represent
the average number of facts and answers within the subject and question, respectively.}
\label{tab:data-stats}
\end{table}
\section{The \cotempqa Datasets}
\subsection{The Taxonomy of Co-temporal Relations}
Co-temporal relations are fundamental to understanding how events interconnect in time.
These relationships highlight when facts or events happen simultaneously, which can be categorized into three distinct types~\cite{pustejovsky2003timeml}, as shown in Table~\ref{tab:allen}.
Each of them represents a unique manner, whether events coincide with or overlap with each other in the temporal aspect.
We divide these relations into four different scenarios below:
\begin{itemize}[leftmargin=*]
\setlength{\itemsep}{0pt}
    \item \textbf{Equal:} Facts occur simultaneously, representing a strict co-temporal relationship, occurring at the same time without duration differences.
    \item \textbf{Overlap:} Facts partially coincide in time. This scenario is common in real-world settings, where facts often intersect for a part of their duration.
    \item \textbf{During:} One fact is entirely contained within the timeline of another, reflecting a more complex interaction of timelines.
    \item \textbf{Mix:} A combination of the three types above. This category is particularly challenging as it involves the complexity and variability of real-world temporal relationships, necessitating a comprehensive level of co-temporal reasoning.
\end{itemize}

\subsection{Structuring Temporal Facts}
We utilize the Wikidata~\cite{Wikidata} dump of September 20, 2023 as our knowledge source for extracting time-dependent facts.
Following \citet{dhingra-etal-2022-time} and \citet{tan-etal-2023-towards}, we focus on nine time-sensitive entity relations \cite{cheng2021hacred,gu2022delving,qu2023distantly} and keep a maximum of 2,000 subjects for each relation type.
To structure the information, we transform the knowledge triples and qualifiers into a quintuplet format of
$(s, r, o, t_{s}, t_{e})$, where $s$ is the subject, $r$ is the relation, $o$ is the object, $t_{s}$ and $t_{e}$ are the start time and end time.
We group all the temporal facts by subject, denoted as $S = \{(s, r_{i}, o_{i}, t_{s_{i}}, t_{e_{i}}) | i \in 1 \dots N \}$, where $N$ is the number of facts within a group. We keep the groups that contain three or more temporal facts. 

\subsection{Extracting Co-temporal Facts}
\label{sec:extract_fact}

\begin{algorithm}[t]
\small
\caption{Identifying Co-temporal Facts}
\label{alg:simplified_fact_analysis_tuple}
\begin{algorithmic}[1]
    \State \textbf{Input:} Set of facts $F$, each fact as $(s, r, o, t_{s}, t_{e})$
    \State \textbf{Output:} Set of co-temporal facts with their minimum temporal units

    \Function{MinMaxTime}{$f_{i}$, $f_{j}$}
        \State $(s_i, r_i, o_i, t_{s_i}, t_{e_i}) \gets f_{i}$
        \State $(s_j, r_j, o_j, t_{s_j}, t_{e_j}) \gets f_{j}$
        \State $start \gets$ max($t_{si}$, $t_{sj}$)
        \State $end \gets$ min($t_{ei}$, $t_{ej}$)
        \State $T_{\text{min}} \gets$ ($start$, $end$)
        \If{$start \leq end$} \Return $T_{\text{min}}$
        \Else \ \Return None
        \EndIf
    \EndFunction

    \State $R \gets$ empty set \Comment{$R$ is the set of co-temporal facts}
    \For{each $f_{i}$ in $F$}
        \For{each $f_{j}$ in $F$ where $f_{i} \neq f_{j}$}
            \State $T_{\text{min}} \gets$ \Call{MinMaxTime}{$f_{i}$, $f_{j}$}
            \If{$T_{\text{min}}$ is not None}
                \State $R \gets R \cup \{ (f_{i}, f_{j}, T_{\text{min}}) \}$
            \EndIf
        \EndFor
    \EndFor

    \State \Return $R$
\end{algorithmic}
\label{alg:min_temporal_unit_facts}
\end{algorithm}

Building on our approach to structuring time-dependent facts from Wikidata, we compare the timestamps of different facts to identify overlaps.
Each fact $f_{i}$ and its co-temporal counterpart $f_{j}$ are represented as a triple, with $f_{i} = \{s_{i}, r_{i}, o_{i}\}$ and $f_{j} = \{s_{j}, r_{j}, o_{j}\}$.
$\mathcal{S} = \{s_{i}, s_{j}\}$,  $\mathcal{R} = \{r_{i}, r_{j}\}$,  $\mathcal{O} = \{o_{i}, o_{j}\}$ are the sets of subjects, relations, and objects within the co-temporal fact pairs $(f_{i}, f_{j})$, respectively.
We categorize fact pairs into five scenarios based on the consistency or variation of $(\mathcal{S}, \mathcal{R}, \mathcal{O})$, involving $(\mathcal{S}, \mathcal{R}, \mathcal{\overline{O}})$, $(\mathcal{S}, \mathcal{\overline{R}}, \mathcal{\overline{O}})$, $(\mathcal{\overline{S}}, \mathcal{R}, \mathcal{O})$, $(\mathcal{\overline{S}}, \mathcal{R}, \mathcal{\overline{O}})$, $(\mathcal{\overline{S}}, \mathcal{\overline{R}}, \mathcal{\overline{O}})$, where an overline indicates a change in the specific set.
For instance, $(\mathcal{S}, \mathcal{R}, \mathcal{\overline{O}})$ represents the scenario where the subjects and relations are constant while the objects differ between $f_{i}$ and $f_{j}$.
We exclude the scenarios $(\mathcal{S}, \mathcal{\overline{R}}, \mathcal{O})$ and $(\mathcal{\overline{S}}, \mathcal{R}, \mathcal{\overline{O}})$ since it is unrealistic for the same subject and object to have different relationships, or for the same object to have the same relationship with different subjects concurrently.
The detailed illustrations are shown in Appendix~\ref{apx:extraction_details}.
Taken $(\mathcal{\overline{S}}, \mathcal{\overline{R}}, \mathcal{\overline{O}})$ as an example, we detail the extraction of co-temporal facts in the \textsc{MinMaxTime} function~(lines 3-11) from Algorithm~\ref{alg:simplified_fact_analysis_tuple}.
This framework identifies the complex co-temporal relations between events, allowing for a more intuitive understanding of how multiple events and states are interrelated in the temporal dimension.

\subsection{QA Pairs Construction}
Upon identifying co-temporal facts $(f_{i}, f_{j}, T_{\text{min}})$, we construct the query $Q$ by the condition fact $f_{\text{cond}}$ and the query fact $f_{\text{query}}$.
$f_{\text{cond}}$ is selected from the intersection fact in $(f_{i}, f_{j})$, while $f_{\text{query}}$ is the other fact in the pair.
To control the correlation between  $f_{\text{cond}}$ and $f_{\text{query}}$, we manually predefined 17 types of relevant relation pairs and constructed questions for the object by these question templates, which can be found in Table~\ref{tab:question-templates}. By predefining these pairs, we align them logically and ensure the extracted facts are contextually interconnected.
Based on the temporal relations identified through $T_{\text{min}}$, we categorize the tasks into four distinct classes: \textbf{Equal}, \textbf{Overlap}, \textbf{During}, and \textbf{Mix}.
Since multiple events can happen simultaneously in real life, we aggregate all valid answers for query $Q$ to a set to avoid questions having several correct answers.
As detailed in Table~\ref{tab:data-stats}, the average number of our answers within the question is 1.42.

\section{The Performance of LLMs on \cotempqa}
\subsection{Experimental Setup}
We investigate the co-temporal reasoning abilities of large language models within two problem settings: 
(1) \textbf{C}losed-\textbf{B}ook \textbf{QA}~(\textbf{CBQA}) is widely recognized task format in time-sensitive QA research~\cite{dhingra-etal-2022-time, liska2022streamingqa, tan-etal-2023-towards}. In this setting, the language model is given only the question and tasked with generating the answer without relying on external natural language texts. The primary challenge here involves the retention and temporal reasoning of knowledge pertinent to the question. (2) In the \textbf{O}pen-\textbf{B}ook \textbf{QA}~(\textbf{OBQA}) setting, we provide all the relevant temporal facts within the group $S = \{(s, r_{i}, o_{i}, t_{s_{i}}, t_{e_{i}}) | i \in 1 ... N \}$ in a structured format directly into the prompt, which is in contrast to previous studies~\cite{chen2021a, wei2023menatqa} that utilized Wikipedia as the knowledge base. This process shifts the evaluation's emphasis towards the reasoning process itself, thereby minimizing the influence of the model's inherent factual extraction capabilities on the outcomes~\cite{tan-etal-2023-towards, chu2023timebench}. Here, the language models need to provide all possible answers within the concurrent timeframe.
\subsection{LLMs for Evaluation}
We perform comprehensive experiments on 14 representative large language models including (1) \textbf{ChatGPT}~\cite{ouyang2022training} ChatGPT is a chat model aligned through Supervised Fine-tuning~(SFT) and Reinforcement Learning from Human Feedback~(RLHF). GPT-4 is an upgraded version of ChatGPT with enhanced reasoning capabilities, making it the most powerful LLM. Since the model is constantly updated, we used the \texttt{gpt3.5-turbo-0613} and \texttt{gpt4-0613} for consistent evaluation. (2) \textbf{LLaMA2}~\cite{touvron2023LLaMA} LLaMA2 is one of the most popular open-source foundation models trained on 2T tokens with efficient group query attention~\cite{ainslie2023gqa}. (3) \textbf{Code-LLaMA}~\cite{roziere2023code}  Code-LLaMA models is a code generation model built on LLaMA2 and further trained
on 500B tokens of code. (4) \textbf{WizardMath}~\cite{luo2023wizardmath} WizardMath is also built on LLaMA2 and further trained on their proposed Reinforcement Learning from Evol-Instruct Feedback~(RLEIF)~\cite{xu2023wizardlm} to enhance the mathematical reasoning abilities of LLaMA2. (5) \textbf{WizardCoder}~\cite{luo2023wizardcoder} WizardCoder, similar to WizardMath, adapts the RLEIF method to
the domain of code. The implementation details of our experiments are shown in Appendix~\ref{apx:implementatio_details}.

\subsection{Evaluation Metrics}
Prior works followed the SQuAD benchmark's evaluation protocol~\cite{rajpurkar-etal-2016-squad}, using exact match~(EM) and token-level $\textrm{F}_1$ score. These metrics calculate the highest scores across all references, tending to overestimate performance in task settings involving questions with multiple possible answers.
Following \citet{zhong2022romqa}, we adopt a stricter \textbf{Acc.} score, where a prediction is correct only if it aligns with all the gold answers for a question. Additionally, we also evaluate our methods by answer-level $\textrm{F}_1$ score~(\textbf{$\textrm{F}_1$}), which is a stricter metric compared to token-level $\textrm{F}_1$ score.
\begin{table*}[!tb]
\begin{small}
    \centering
    \small
    \resizebox{\linewidth}{!}{
    \begin{tabular}{lcccccccccccc|c}
         \toprule
        \multirow{2}{*}{\bf Model} &
        \multicolumn{3}{c}{\textbf{Equal}} & \multicolumn{3}{c}{\textbf{Overlap}} & \multicolumn{3}{c}{\textbf{During}} & \multicolumn{3}{c|}{\textbf{Mix}} & \multirow{2}{*}{\textbf{Overall}} \\
        \cmidrule(lr){2-4} \cmidrule(lr){5-7} \cmidrule(lr){8-10} \cmidrule(lr){11-13}
        & \bf Acc. & \bf $\textrm{F}_1$ & \bf Avg. & \bf Acc. & \bf $\textrm{F}_1$ & \bf Avg. & \bf Acc. & \bf $\textrm{F}_1$ & \bf Avg. & \bf Acc. & \bf $\textrm{F}_1$ & \bf Avg. \\
        \specialrule{0.06em}{2pt}{2pt}
        \multicolumn{14}{c}{\bf{{The Closed Book Question Answer~(CBQA) setting}}} \\
        \textsc{GPT-3.5-Turbo} & \bf 13.8\textcolor{white}{0} & \bf 14.8\textcolor{white}{0} & \bf 14.3\textcolor{white}{0} & 11.3\textcolor{white}{0} & \bf 14.3\textcolor{white}{0} & 12.8\textcolor{white}{0} & \bf 15.0\textcolor{white}{0} & \bf 22.9\textcolor{white}{0} & \bf 18.9\textcolor{white}{0} & \bf 0.0\textcolor{white}{0} & \bf 15.5\textcolor{white}{0} & \bf 7.7\textcolor{white}{0} &\bf 16.3\textcolor{white}{0}\\
        \textsc{GPT-4} & 11.2\textcolor{white}{0} & 12.3\textcolor{white}{0} & 11.8\textcolor{white}{0} & \bf 11.5\textcolor{white}{0} & 14.0\textcolor{white}{0} & \bf 12.7\textcolor{white}{0} & 14.8\textcolor{white}{0} & 18.5\textcolor{white}{0} & 16.7\textcolor{white}{0} & \bf 0.0\textcolor{white}{0} & 13.6\textcolor{white}{0} & 6.8\textcolor{white}{0} & 14.5\textcolor{white}{0}\\     
        \specialrule{0.06em}{2pt}{2pt}
 
        \multicolumn{14}{c}{\bf{{The Open Book Question Answer~(OBQA) setting}}} \\
        \textsc{GPT-3.5-Turbo} & 59.4\textcolor{white}{0} & 66.3\textcolor{white}{0} & 62.8\textcolor{white}{0} & 40.1\textcolor{white}{0} & 48.5\textcolor{white}{0} & 44.3\textcolor{white}{0} & 31.5\textcolor{white}{0} & 42.9\textcolor{white}{0} & 37.2\textcolor{white}{0} & 0.7\textcolor{white}{0} & 46.1\textcolor{white}{0} & 23.4\textcolor{white}{0} & 38.9\textcolor{white}{0}\\
        \textsc{GPT-4} & \bf 91.1\textcolor{white}{0} & \bf 94.3\textcolor{white}{0} & \bf 92.7\textcolor{white}{0} & \bf 55.3\textcolor{white}{0} & \bf 63.5\textcolor{white}{0} & \bf 59.4\textcolor{white}{0} & \bf 44.3\textcolor{white}{0} & \bf 55.8\textcolor{white}{0} & \bf 50.1\textcolor{white}{0} & \bf 23.4\textcolor{white}{0} &\bf 66.5\textcolor{white}{0} &\bf 45.0\textcolor{white}{0} &\bf 54.7\textcolor{white}{0} \\       
        \specialrule{0.04em}{2pt}{2pt}
        \textsc{CodeLLaMA-7B} & 6.4\textcolor{white}{0} & \bf 27.7\textcolor{white}{0} & \bf 17.0\textcolor{white}{0} & 3.1\textcolor{white}{0} & 14.6\textcolor{white}{0} & 8.8\textcolor{white}{0} & 3.1\textcolor{white}{0} & 15.8\textcolor{white}{0} & 9.5\textcolor{white}{0} & \bf 2.0\textcolor{white}{0} & \bf 24.1\textcolor{white}{0} &  \bf 13.0\textcolor{white}{0} & 10.5\textcolor{white}{0} \\
        \textsc{WizardCoder-7B} & 9.2\textcolor{white}{0} & 21.1\textcolor{white}{0} & 15.1\textcolor{white}{0} &  4.7\textcolor{white}{0} & 14.8\textcolor{white}{0} & 9.8\textcolor{white}{0} & 6.3\textcolor{white}{0} &  15.9\textcolor{white}{0} & 11.1\textcolor{white}{0} & 0.5\textcolor{white}{0} & 20.4\textcolor{white}{0} & 10.5\textcolor{white}{0} & 11.2\textcolor{white}{0}  \\
        \textsc{LLaMA-7B} & 4.1\textcolor{white}{0} & 18.9\textcolor{white}{0} & 11.5\textcolor{white}{0} & 4.7\textcolor{white}{0} & \bf 19.5\textcolor{white}{0} & 12.1\textcolor{white}{0} & 4.5\textcolor{white}{0} & 19.5\textcolor{white}{0} & 12.0\textcolor{white}{0} & 0.2\textcolor{white}{0} & 23.8\textcolor{white}{0} & 12.0\textcolor{white}{0} & 12.0\textcolor{white}{0}\\
        \textsc{WizardMath-7B} & \textbf{12.4}\textcolor{white}{0} & \text{16.5}\textcolor{white}{0} & \text{14.4}\textcolor{white}{0} & \textbf{9.2}\textcolor{white}{0} & 15.2\textcolor{white}{0} & \bf 12.2\textcolor{white}{0} & \textbf{11.6}\textcolor{white}{0} & \textbf{20.5}\textcolor{white}{0} & \bf 16.0\textcolor{white}{0} & 0.4\textcolor{white}{0} & 22.0\textcolor{white}{0} & 11.2\textcolor{white}{0} & \bf 14.8\textcolor{white}{0}  \\
        \specialrule{0.04em}{1pt}{1pt}
        \textsc{CodeLLaMA-13B} & \text{7.6}\textcolor{white}{0} & \text{28.3}\textcolor{white}{0} & \text{18.0}\textcolor{white}{0} & \text{4.1}\textcolor{white}{0} & 17.0\textcolor{white}{0} & 10.6\textcolor{white}{0} & \text{3.3}\textcolor{white}{0} & \text{19.3}\textcolor{white}{0} & 11.3\textcolor{white}{0} & \bf 3.2\textcolor{white}{0} & \bf 28.6\textcolor{white}{0} & \bf 15.9\textcolor{white}{0} & 12.4\textcolor{white}{0}  \\
        \textsc{WizardCoder-13B} & 8.3\textcolor{white}{0} & 16.6\textcolor{white}{0} & 12.4\textcolor{white}{0} &  7.0\textcolor{white}{0} & 17.8\textcolor{white}{0} & 12.4\textcolor{white}{0} & 9.5\textcolor{white}{0} &  19.7\textcolor{white}{0} & \bf 14.6\textcolor{white}{0} & 1.1\textcolor{white}{0} & 24.1\textcolor{white}{0} & 12.6\textcolor{white}{0} & 13.9\textcolor{white}{0}  \\  
        \textsc{LLaMA-13B} & 11.2\textcolor{white}{0} & \bf 31.2\textcolor{white}{0} & 21.2\textcolor{white}{0} & 5.8\textcolor{white}{0} & \bf 21.6\textcolor{white}{0} & \bf 13.7\textcolor{white}{0} & 5.0\textcolor{white}{0} &  \bf 20.6\textcolor{white}{0} & 12.8\textcolor{white}{0} & 1.1\textcolor{white}{0} & 26.9\textcolor{white}{0} & 14.0\textcolor{white}{0} & 13.8\textcolor{white}{0}\\
        \textsc{WizardMath-13B} & \textbf{23.9}\textcolor{white}{0} & \text{29.0}\textcolor{white}{0} & \textbf{26.4}\textcolor{white}{0} & \textbf{10.9}\textcolor{white}{0} & 15.1\textcolor{white}{0} & 13.0\textcolor{white}{0} & \textbf{11.7}\textcolor{white}{0} & \text{17.1}\textcolor{white}{0} & 14.4\textcolor{white}{0} & 0.0\textcolor{white}{0} & 13.2\textcolor{white}{0} & 6.6\textcolor{white}{0} & \bf 14.4\textcolor{white}{0}  \\
        \specialrule{0.04em}{1pt}{1pt}
        \textsc{CodeLLaMA-34B} & \text{16.1}\textcolor{white}{0} & \text{46.5}\textcolor{white}{0} & \text{31.3}\textcolor{white}{0} & \text{9.8}\textcolor{white}{0} & 27.0\textcolor{white}{0} & 18.4\textcolor{white}{0} & \text{8.1}\textcolor{white}{0} & \text{28.4}\textcolor{white}{0} & 18.3\textcolor{white}{0} & 4.4\textcolor{white}{0} & 40.3\textcolor{white}{0} & 22.4\textcolor{white}{0} & 20.0\textcolor{white}{0}   \\ 
        \textsc{WizardCoder-34B} & 19.5\textcolor{white}{0} & 26.3\textcolor{white}{0} & 22.9\textcolor{white}{0} &  15.2\textcolor{white}{0} & 22.4\textcolor{white}{0} & 18.8\textcolor{white}{0} & 15.9\textcolor{white}{0} &  23.9\textcolor{white}{0} & 19.9\textcolor{white}{0} & 0.9\textcolor{white}{0} & 25.9\textcolor{white}{0} & 13.4\textcolor{white}{0} & 19.2\textcolor{white}{0}  \\
        \textsc{LLaMA-70B} & \text{11.9}\textcolor{white}{0} & \text{41.7}\textcolor{white}{0} & \text{26.8}\textcolor{white}{0} & \text{10.0}\textcolor{white}{0} & 32.5\textcolor{white}{0} & 21.2\textcolor{white}{0} & \text{9.4}\textcolor{white}{0} & \text{33.5}\textcolor{white}{0} & 21.4\textcolor{white}{0} & \bf 5.2 \textcolor{white}{0} & \bf 42.5\textcolor{white}{0} & \bf 23.8\textcolor{white}{0} & 22.2\textcolor{white}{0}   \\ 
        \textsc{WizardMath-70B} & \bf 36.7\textcolor{white}{0} & \bf 46.8\textcolor{white}{0} & \bf 41.8\textcolor{white}{0} & \bf 23.6\textcolor{white}{0} & \bf 33.7\textcolor{white}{0} & \bf 28.6\textcolor{white}{0} & \bf 25.5\textcolor{white}{0} & \bf 37.1\textcolor{white}{0} & \bf 31.3\textcolor{white}{0} & 0.4\textcolor{white}{0} & 32.9\textcolor{white}{0} & 16.6\textcolor{white}{0} & \bf 30.1\textcolor{white}{0}  \\
        \specialrule{0.04em}{1pt}{1pt}
        \textsc{Human} & 97.0\textcolor{white}{0} & 98.3\textcolor{white}{0} & 97.7\textcolor{white}{0} &  91.1\textcolor{white}{0} & 93.5\textcolor{white}{0} & 92.3\textcolor{white}{0} & 82.0\textcolor{white}{0} &  87.0\textcolor{white}{0} & 84.5\textcolor{white}{0} & 88.0\textcolor{white}{0} & 96.2\textcolor{white}{0} & 92.1\textcolor{white}{0} & 92.8\textcolor{white}{0}  \\

        \bottomrule
    \end{tabular}}
    \caption{Experimental results of each model in the \textbf{CBQA} and \textbf{OBQA} settings of our proposed \cotempqa. Notably, we only report the performance of GPT-3.5 and GPT-4 in \textbf{CBQA} setting as the open-source LLMs are almost negligible here, and closed-book human evaluations largely depend on individual knowledge, leading to significant variations between different individuals. The best performance of each model is in \textbf{bold}.}
    \label{table:main_table}
\end{small}
\end{table*}

\subsection{Results and Analysis}
The main results are shown in Table~\ref{table:main_table}. We report human performance to serve as an upper bound. From the results, we can observe:

\paragraph{LLMs partially grasp co-temporal reasoning}
Our analysis reveals that, despite GPT-4 exhibiting the best performance among all LLMs, there is still a considerable disparity compared to human performance~(54.7 vs. 92.8), indicating significant potential for further improvement in co-temporal reasoning.
We also discover that models exhibit different reasoning capabilities in different co-temporal scenarios.
Take GPT-4 for further illustration, in the simple co-temporal reasoning task, i.e., the \textbf{Equal} scenario, GPT-4 demonstrates strong performance, achieving a 92.7 score overall.
However, its performance significantly declines in more complex scenarios.
Specifically, in the \textbf{Overlap} category, GPT-4's accuracy falls to 59.4, decreasing further to 50.1 in the \textbf{During} category. In the most challenging category, \textbf{Mix}, which combines various temporal relations, GPT-4's performance drops to 45.0.
We provide a case study to explain the varying difficulties of scenarios and model performances in Appendix~\ref{apx:case_study}.
As shown in Table~\ref{tb:case-study}, the concurrent characteristics in the \textbf{Equal} scenario are relatively obvious compared with 
 the \textbf{Overlap} and \textbf{During} scenarios.
 Furthermore, the \textbf{Mix} scenario has more than one answer and involves reasoning with multiple co-temporal relations, which makes it the most challenging compared to other scenarios.

\paragraph{CBQA is more challenging for LLMs}
LLMs exhibit significantly weaker performance in the \textbf{CBQA} compared to the \textbf{OBQA}, as reflected in the GPT-4's performance~(14.5 vs. 54.7).
Interestingly, GPT-4 is outperformed by GPT-3.5 in \textbf{CBQA}. Our error analysis indicates that GPT-4 often responds with \textit{``uncertain"} when unsure, unlike GPT-3.5, which tends to provide direct answers. This discovery is also found in previous works~\cite{openai2023gpt4, wei2023menatqa}.
This characteristic hinders GPT-4's effectiveness in co-temporal \textbf{CBQA}, where precise answers are needed.
While constructing our datasets, we concentrated on the top 2,000 subjects for each relationship type. These subjects are typically well-covered in pre-training stages, as Wikipedia is a significant part of their training data~\cite{touvron2023LLaMA}.
Despite this prior exposure, LLMs' reduced capability in \textbf{CBQA} underscores the need to enhance the co-temporal reasoning abilities of LLMs, empowering them to comprehend and reason about concurrent events.

\begin{figure*}
\begin{subfigure}[b]{0.5\textwidth}
\centering
\includegraphics[width=\textwidth]{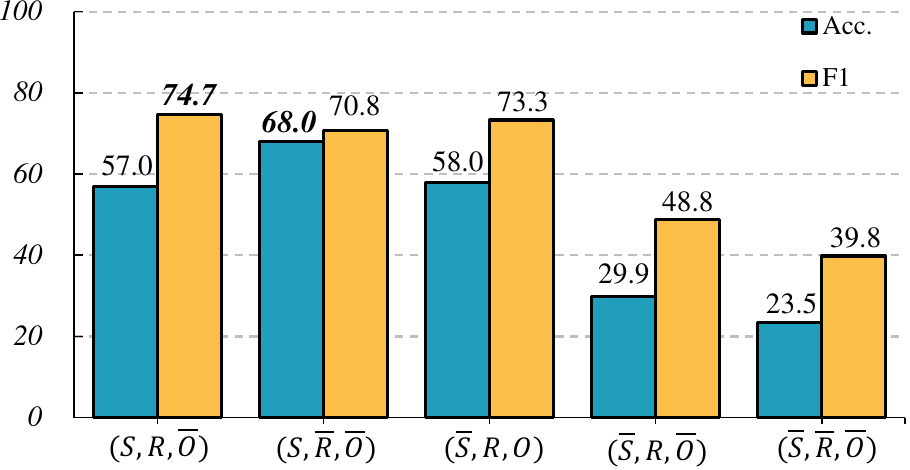}
\caption{Triple Element Types}
\label{fig:SRO}
\end{subfigure}
\hfill
\begin{subfigure}[b]{0.5\textwidth}
\centering
\includegraphics[width=\textwidth]{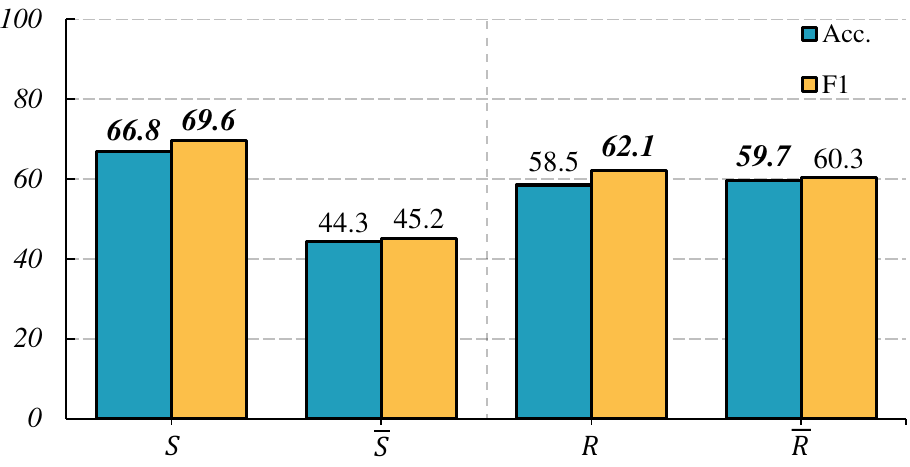}
\caption{Triplet Element Types}
\label{fig:SR}
\end{subfigure}
\caption{Performance of GPT-4 under different co-temporal element types in the \textbf{OBQA} setting of our \cotempqa. \textit{``Triplet''} indicates scenarios where each element of $(\mathcal{S}$, $\mathcal{R}$, $\mathcal{O})$ either changes or remains constant, while \textit{``Triple''} focuses on variations in a single element. The overline indicates we changed the element in fact to others. The best performance of each element type is \textbf{bold}.}
\label{fig:SRO&SR}
\end{figure*}

\paragraph{Different aspects of capability benefit co-temporal reasoning differently}
Notably, models specialized in mathematical reasoning~(e.g., WizardMath-70B) show significant improvements in co-temporal reasoning, scoring 30.1, compared to the foundational LLaMA-70B model's 22.2 and CodeLLaMA-34B's 20.0.
This improvement indicates a strong correlation between the skills utilized in math and those required for understanding and interpreting complex temporal relationships.  
Although WizardMath is the overall best model among the baseline, we also observe its reduced effectiveness in the \textbf{Mix} scenario compared with others.
By further investigation, questions have multiple answers in the \textbf{Mix} scenario.
WizardMath tends to return a single response rather than enumerating all possible answers, causing higher precision but lower recall in contrast to other models~(i.e., LLaMA, CodeLLaMA).
We provide further experimental results and analysis in Appendix~\ref{apx:mix_analysis}.

In Appendix~\ref{apx:error_analysis}, we provide a detailed error analysis to help understand the limitations of current models and guide future improvements in co-temporal reasoning capabilities.

\begin{table*}[!tb]
\begin{small}
    \centering
    \small
    \resizebox{\linewidth}{!}{
    \begin{tabular}{lcccccccccccc|c}
         \toprule
        \multirow{2}{*}{\bf Model} &
        \multicolumn{3}{c}{\textbf{Equal}} & \multicolumn{3}{c}{\textbf{Overlap}} & \multicolumn{3}{c}{\textbf{During}} & \multicolumn{3}{c|}{\textbf{Mix}} & \multirow{2}{*}{\textbf{Overall}} \\
        \cmidrule(lr){2-4} \cmidrule(lr){5-7} \cmidrule(lr){8-10} \cmidrule(lr){11-13}
        & \bf Acc. & \bf $\textrm{F}_1$ & \bf Avg. & \bf Acc. & \bf $\textrm{F}_1$ & \bf Avg. & \bf Acc. & \bf $\textrm{F}_1$ & \bf Avg. & \bf Acc. & \bf $\textrm{F}_1$ & \bf Avg. \\
        \specialrule{0.06em}{2pt}{2pt}
        \multicolumn{14}{c}{\bf{{The Closed Book Question Answer~(CBQA) setting}}} \\
        \textsc{GPT-4} & 11.2\textcolor{white}{0} & 12.3\textcolor{white}{0} & 11.8\textcolor{white}{0} & 11.5\textcolor{white}{0} & 14.0\textcolor{white}{0} & 12.7\textcolor{white}{0} & 14.8\textcolor{white}{0} & 18.5\textcolor{white}{0} & 16.7\textcolor{white}{0} & 0.0\textcolor{white}{0} & 13.6\textcolor{white}{0} & 6.8\textcolor{white}{0} & 14.5\textcolor{white}{0}\\   
        \textsc{+ CoT} & 12.2 \textcolor{white}{0} & 14.4\textcolor{white}{0} & 13.3\textcolor{white}{0} & 8.4\textcolor{white}{0} & 12.5\textcolor{white}{0} & 10.5\textcolor{white}{0} & 12.1\textcolor{white}{0} & 18.6\textcolor{white}{0} &  15.3\textcolor{white}{0} & 1.6\textcolor{white}{0} & 14.3\textcolor{white}{0} & 8.0\textcolor{white}{0}  &  13.6\textcolor{white}{0} \\
        \textsc{+ Fs} &  26.4\textcolor{white}{0} & 29.6\textcolor{white}{0} &  {28.0}\textcolor{white}{0} & 17.6\textcolor{white}{0} &  21.2\textcolor{white}{0} & 19.4\textcolor{white}{0} & 20.6\textcolor{white}{0} & 26.7\textcolor{white}{0} & 23.7\textcolor{white}{0} & 0.0\textcolor{white}{0} & 21.7\textcolor{white}{0} & 10.9\textcolor{white}{0} & 22.0\textcolor{white}{0}\\    
        \textsc{+ Fs\&CoT} & \bf 32.1\textcolor{white}{0} & \bf 35.2\textcolor{white}{0} & \bf 33.6\textcolor{white}{0} &\bf 19.9\textcolor{white}{0} & 25.7\textcolor{white}{0} &  22.8\textcolor{white}{0} & \bf 23.2\textcolor{white}{0} & 29.5\textcolor{white}{0} &26.4\textcolor{white}{0} & 0.5\textcolor{white}{0} & 25.6\textcolor{white}{0} & 13.1\textcolor{white}{0} & 25.0\textcolor{white}{0}\\  
        \textsc{+ Fs\&Mr-CoT} & 24.8\textcolor{white}{0} & 30.6\textcolor{white}{0} &  27.7\textcolor{white}{0} &16.7\textcolor{white}{0} &  \bf 29.9\textcolor{white}{0} & \bf 23.3\textcolor{white}{0} & 20.8\textcolor{white}{0} & \bf 35.7\textcolor{white}{0} &\bf 28.2\textcolor{white}{0} & \bf 3.9\textcolor{white}{0} &\bf 31.7\textcolor{white}{0} &\bf 17.8\textcolor{white}{0} & \bf 26.3\textcolor{white}{0}\\  
        \specialrule{0.06em}{2pt}{2pt}
        \multicolumn{14}{c}{\bf{{The Open Book Question Answer~(OBQA) setting}}} \\
        \textsc{GPT-4} & 91.1\textcolor{white}{0} & 94.3\textcolor{white}{0} & 92.7\textcolor{white}{0} & 55.3\textcolor{white}{0} & 63.5\textcolor{white}{0} & 59.4\textcolor{white}{0} & 44.3\textcolor{white}{0} & 55.8\textcolor{white}{0} & 50.1\textcolor{white}{0} & 23.4\textcolor{white}{0} & 66.5\textcolor{white}{0} & 45.0\textcolor{white}{0} & 54.7\textcolor{white}{0} \\   
        \textsc{+ CoT} & 87.8 \textcolor{white}{0} & 90.0\textcolor{white}{0} & 88.9\textcolor{white}{0} & 46.2\textcolor{white}{0} & 58.7\textcolor{white}{0} & 52.5\textcolor{white}{0} & 43.5\textcolor{white}{0} & 57.0\textcolor{white}{0} &  50.2\textcolor{white}{0} & 29.5\textcolor{white}{0} & 71.6\textcolor{white}{0} & 50.5\textcolor{white}{0} & 54.1\textcolor{white}{0} \\
        \textsc{+ Fs} & 87.4\textcolor{white}{0} & 91.4\textcolor{white}{0} &  89.4\textcolor{white}{0} & 62.6\textcolor{white}{0} & 72.5\textcolor{white}{0} & 67.6\textcolor{white}{0} & 55.9\textcolor{white}{0} & 68.6\textcolor{white}{0} & 62.2\textcolor{white}{0} & 30.6\textcolor{white}{0} & 71.9\textcolor{white}{0} & 51.2\textcolor{white}{0}& 64.2\textcolor{white}{0}\\    
        \textsc{+ Fs\&CoT} & \bf 96.8\textcolor{white}{0} & \bf 97.3\textcolor{white}{0} & \bf 97.1\textcolor{white}{0} & 61.3\textcolor{white}{0} & 71.4\textcolor{white}{0} & 66.3\textcolor{white}{0} &  55.7\textcolor{white}{0} & 69.4\textcolor{white}{0} & 62.5\textcolor{white}{0} &32.1\textcolor{white}{0} &73.2\textcolor{white}{0} &52.7\textcolor{white}{0} & 65.0\textcolor{white}{0}\\          
        \textsc{+ Fs\&Mr-CoT} & 95.9\textcolor{white}{0} & 97.2\textcolor{white}{0} & 96.5\textcolor{white}{0} & \bf 77.9\textcolor{white}{0} & \bf 83.9\textcolor{white}{0} & \bf 80.9\textcolor{white}{0} & \bf 69.0\textcolor{white}{0} & \bf 78.8\textcolor{white}{0} & \bf 73.9\textcolor{white}{0} &\bf 50.3\textcolor{white}{0} &\bf 82.2\textcolor{white}{0}  &\bf 66.2\textcolor{white}{0} & \bf 75.8\textcolor{white}{0}\\      

        \bottomrule
    \end{tabular}}
    \caption{Performance of GPT-4 under Zero-shot CoT~(\textsc{CoT}) prompting, Few-shot~(\textsc{Fs}) prompting, Few-shot CoT~(\textsc{Fs\&CoT}) prompting and our proposed Few-Shot Mr-CoT~(\textsc{FS\&Mr-CoT}) prompting in \textbf{CBQA} and \textbf{OBQA}.}
    \label{table:prompt_table}
\end{small}
\end{table*}

\begin{table*}[t!]
\resizebox{\linewidth}{!}{
    \centering
\begin{tabular}{ll}
\toprule
\textbf{Question}               & 
\begin{tabular}[c]{@{}l@{}}While Klaus Fuchs was working for Atomic Energy Research Establishment, \\which employer did Alexander Oppenheim work for during the identical time period?  
\end{tabular}                                                                                                                                                                                                                                                                                                                                                                              \\ \hline
\textbf{Context}                & 
\begin{tabular}[c]{@{}l@{}}
    Alexander Oppenheim works for National University of Singapore from 1949 to 1965. 
    \\ Alexander Oppenheim works for University of Malaya from 1949 to 1965.
    \\ Klaus Fuchs works for Atomic Energy Research Establishment from 1946 to 1950.
    \\ ......
\end{tabular} \\ \hline
\textbf{Gold Answer}            & \cellcolor[HTML]{FFFFFF}\textcolor{blue}{\textbf{National University of Singapore and University of Malaya}}                                                                                                                                                                                                                                                                                                                            \\ \hline
\textbf{GPT-4} & Answer: \cellcolor[HTML]{FFFFFF}\textcolor{red}{\textbf{National University of Singapore}}                                                                                                                                                                                                                                                                                                                                     \\ \hline
\textbf{+ CoT}      & 
\begin{tabular}[c]{@{}l@{}}
  Reasoning: Klaus Fuchs worked for Atomic Energy Research Establishment from 1946 to 1950. 
    \\ Currently, Alexander Oppenheim worked for University of Ghana from 1968 to 1973.
    \\However, this does not overlap with the time Klaus Fuchs was at the Atomic Energy Research Establishment. \\
    Answer: \textcolor{red}{\textbf{no information}} 
  
\end{tabular}                                                                                                                                                                                                   \\ \hline
\textbf{+ Fs}       & Answer: \cellcolor[HTML]{FFFFFF}\textcolor{red}{\textbf{University of Malaya}}                                                                                                                                                                                                                                                                                                                                     \\ \hline
\textbf{+ Fs\&CoT}            & 
\begin{tabular}[c]{@{}l@{}} Reasoning: Klaus Fuchs 
    worked for Atomic Energy Research Establishment from 1946 to 1950. 
    \\During this same period, Alexander Oppenheim worked for National University of Singapore, \\as his tenure with the university spanned from 1949 to 1965.\\
    Answer: \textcolor{red}{\textbf{National University of Singapore}} 
\end{tabular}                                                                                                                                                                                                                                             \\ \hline

\textbf{+ Fs\&Mr-CoT} & 
\begin{tabular}[c]{@{}l@{}}
    The datetime can be formed as ((1946, None, None), (1950, None, None)). \\
    The content provided and related to the question can be structured as: \\
    (University of Malaya, (1949, None, None), (1965, None, None)) \\
    (National University of Singapore, (1949, None, None), (1965, None, None)) \\
    Given the ((1946, None, None), (1950, None, None)), compared with all contents related, \\
    we find that $[(1949, \text{None}, \text{None}) - (1965, \text{None}, \text{None})] \cap [(1946, \text{None}, \text{None}) - (1950, \text{None}, \text{None})] \neq \emptyset$ \\ Reasoning: Klaus Fuchs worked for Atomic Energy Research Establishment from 1946 to 1950.\\
    Answer: \textcolor{blue}{\textbf{National University of Singapore and University of Malaya}} 
\end{tabular} \\
\bottomrule
\end{tabular}}
    \caption{Example inputs and outputs of GPT-4 with Zero-shot CoT~(\textsc{CoT}) prompting, Few-shot~(\textsc{Fs}) prompting, Few-shot CoT~(\textsc{FS\&CoT}) prompting and Few-Shot Mr-CoT~(\textsc{FS\&Mr-CoT}). Answers highlighted in blue are correct, whereas thoses marked in red are incorrect.}
    \label{tb:cot}
\end{table*}

\subsection{Data Analysis}
In Section~\ref{sec:extract_fact}, we categorize co-temporal facts into five scenarios.
Building on this classification, this section  delves into investigating how various types of fact elements influence LLMs' ability to perform co-temporal reasoning.
To ensure fairness in our experiments, we excluded questions with multiple answers and standardized the number of questions across all co-temporal relations.
Figure~\ref{fig:SRO&SR} illustrates GPT-4's performance with various element types. Additional results concerning different LLMs are presented in Table~\ref{table:extensive_SRO}, and results consistently align with the findings shown below:

\paragraph{The influence of triple element types} As observed in Figure~\ref{fig:SRO}, the complexity of co-temporal reasoning for models increases with the number of changing elements.
Among the scenarios, $(\mathcal{\overline{S}}, \mathcal{R}, \mathcal{\overline{O}})$, $(\mathcal{\overline{S}}, \mathcal{\overline{R}}, \mathcal{\overline{O}})$ are particularly challenging compared to others.
It indicates that LLMs encounter significant challenges when dealing with scenarios of high complexity, where multiple elements undergo simultaneous changes.
The analysis below further investigates which elements present the most significant challenges to co-temporal reasoning capabilities.

\paragraph{The influence of triplet element types} 
In the left part of Figure~\ref{fig:SR}, we observe a notable decline in the model's performance, i.e., 22.5 point decrease in \textbf{Acc.} and 24.4 in \textbf{$\textrm{F}_1$} when it engages in reasoning involving multiple subjects.
These findings highlight models' inherent difficulty when processing information from multiple concurrent subjects.
While the reasoning process for handling multiple subjects shares similarities with single-subject scenarios, real-world situations are inherently more complex and variable. 
The model is required to integrate information across different subjects and understand complex relationships that extend beyond a single domain or context.
On the other hand, in the right part of Figure~\ref{fig:SR}, we discover that the difference in the fact relation does not significantly impact the abilities of LLMs in co-temporal reasoning.
This is attributed to the fact that relationship changes are not as complex as those involving multiple subjects in real-world scenarios~\cite{huang2023reasoning}, making them less challenging for the models' capabilities.

\section{Making Language Models Better \cotempqa Responders}

Previous research has demonstrated that the Chain-of-Thought (CoT) enables models to process complex reasoning tasks, such as mathematical and logical reasoning, in a step-by-step manner~\cite{NEURIPS2022_9d560961}.
Motivated by this, we explore the application of CoT prompting to improve the capabilities of co-temporal reasoning in this section.

\subsection{Math-reasoning CoT~(\textsc{Mr-CoT})}
As indicated in Table~\ref{table:main_table}, our research uncovers a mathematically oriented reasoning framework that can enrich the LLMs' understanding and handling of co-temporal reasoning tasks.
Specifically, the WizardMath-70B model scores 30.1 overall, notably higher than the foundational LLaMA-70B model's score of 22.2.
In light of this finding, we propose a \textsc{Math-reasoning}~(\textsc{Mr-CoT}) instruction-based prompting, which can be used together with in-context learning and chain-of-thought prompting.
As demonstrated in the bottom of Table~\ref{tb:cot}, our framework consists of three steps: (1) establish the key datetime, (2) structure the relevant timeline, and (3) mathematically identify the overlap.
This prompt aims to guide the LLMs towards approaching temporal reasoning problems through a mathematical perspective, aligning their problem-solving processes more closely with mathematical logic and principles.

\subsection{Experimental Setup}
We launch experiments under both zero-shot and few-shot settings. In the zero-shot CoT scenario, we use \textit{Let’s think step by step}~\cite{NEURIPS2022_8bb0d291} after questions as the reasoning trigger.
In contrast, the few-shot setting provides the model with several question-answer pairs as initial demonstrations.
Specifically, for the few-shot CoT scenario, we manually create rationales for each task, which are used as demonstrations to guide the model in step-by-step reasoning. Further details on the instructions and demonstrations are available from Figure~\ref{fig:start} to Figure~\ref{fig:end} in Appendix~\ref{apx:prompt}.

\subsection{Results and Analysis}
The results are presented in Table~\ref{table:prompt_table}, and the output of GPT-4 to a range of prompts under different settings are shown in Table~\ref{tb:cot}. From these tables, we can discover the following insights:
\paragraph{Inconsistency in the impact of existing CoT prompts on GPT-4}
In the zero-shot scenario, improvements were inconsistent, with a notable 5.5 performance increase in the \textbf{Mix} task and a 3.8 decrease in the \textbf{Equal} task under the \textbf{OBQA} setting. This suggests that the impact of CoT prompts varies significantly based on the task type.
Moreover, GPT-4 demonstrates an overall decline in performance on both \textbf{CBQA} and \textbf{OBQA} when complemented with CoT.
In the few-shot scenario, while overall improvements exist due to CoT prompts, these are relatively modest, amounting to an average performance enhancement of 0.8 in \textbf{OBQA}. All results indicate that while existing CoT prompts can be beneficial, their effectiveness is nuanced and task-dependent.

\paragraph{Superiority of our proposed \textsc{Mr-CoT}}
Our method demonstrates significant superiority over existing reasoning enhancement strategies. Notably, \textsc{Mr-CoT} significantly enhances performance on the more challenging tasks, yielding improvements of 14.6, 11.4, and 13.5 on the tasks \textbf{Overlap}, \textbf{During}, and \textbf{Mix}, respectively in the \textbf{OBQA} setting. In the closed-book scenario, which is typically more challenging to improve, our method still achieves a 1.3 enhancement. However, it is observed that our method has a moderate effect on the \textbf{Equal} setting. We hypothesize that this is because this task is simple enough and does not require the additional complexity of mathematical reasoning. In such cases, this added complexity could be counterproductive.
Despite these advancements, there is still a considerable gap compared to human-level reasoning, indicating the need for more effective methods to improve the model's co-temporal reasoning abilities.

\section{Related Work}
\subsection{Temporal Reasoning Benchmarks}
Temporal reasoning in natural language processing has seen significant advancements over the years. Early benchmarks, such as TimeBank~\cite{pustejovsky2003timebank}, and TempEval-3~\cite{DBLP:journals/corr/abs-1206-5333}, lay the foundational work in this domain. They primarily focused on understanding temporal relationships between events in text, offering a preliminary framework for analyzing time in language models. 
However, recent years have witnessed a significant surge in developing time-sensitive question-answering datasets. These newer datasets, including MC-TACO~\cite{zhou-etal-2019-going}, SituatedQA~\cite{zhang-choi-2021-situatedqa}, TimeQA~\cite{chen2021a}, TempLAMA~\cite{dhingra-etal-2022-time}, StreamingQA~\cite{liska2022streamingqa}, RealtimeQA~\cite{kasai2022realtime}, TempREASON~\cite{tan-etal-2023-towards} and Menatqa~\cite{wei2023menatqa}, represent a more nuanced approach to temporal reasoning. These datasets challenge models to answer questions grounded in specific times or events, thereby testing the models' ability to comprehend and reason with temporal information more dynamically.
The introduction of benchmarks such as TRAM~\cite{wang2023tram} and TimeBench~\cite{chu2023timebench} marks a significant advancement, providing crucial platforms for temporal reasoning research.
Despite these advancements, there has been a noticeable gap in exploring the concurrent nature of temporal events. Previous research has primarily focused on individual events or sequences of events in isolation, overlooking the complexity of scenarios where multiple events co-occur or interact over the same period.
Our work aims to fill this gap by being the first to explore the concurrent nature of temporal events.

\subsection{Temporal Reasoning over LLMs}
To enhance the temporal reasoning capabilities of language models, previous methods either rely heavily on knowledge graphs to rank entities that satisfy the time-related queries~\cite{han-etal-2021-econet, Mavromatis_Subramanyam_Ioannidis_Adeshina_Howard_Grinberg_Hakim_Karypis_2022, liang2022reasoning, chen-etal-2023-multi, huang2024joint} or are strictly dependent on the continual pre-training to strengthen models' abilities in certain temporal aspects~\cite{su-etal-2022-improving, tan-etal-2023-towards, yuan2023future}.
The evolution of LLMs has demonstrated impressive ability in complex reasoning tasks~\cite{chen-2023-large}, such as mathematical reasoning~\cite{mishra-etal-2022-lila} and logic reasoning~\cite{liu2023glore}.
In light of these advancements, recent methods shift towards a program-aided approach~\cite{gao2023pal} to improve the performance of time-sensitive tasks, employing Python code as an intermediate logical step instead of natural language~\cite{li2023unlocking}. This method, while effective, relies heavily on external tools~\cite{zhu2023question} and does not fully leverage the inherent capabilities of LLMs~\cite{brown2020language}.
The results reveal that existing LLMs, even with advanced strategies like Chain of Thought~\cite{NEURIPS2022_9d560961}, demonstrate limited efficacy in addressing the complexities inherent in co-temporal reasoning tasks.
Meanwhile, our research highlights the significant role of mathematical abilities in co-temporal reasoning, offering a direction for future methodologies.

\section{Potential impact}
We believe that it is crucial to enable LLMs to understand events that occur simultaneously or overlap in time. The potential impact of improved co-temporal reasoning on downstream applications includes the following:
\begin{itemize}[leftmargin=*]
\setlength{\itemsep}{0pt}
\item \textbf{Question Answering:} Advanced co-temporal reasoning capabilities enable models to better handle questions involving concurrent events, leading to more accurate responses that reflect a deeper understanding of temporal relationships between events.
\item \textbf{Event Understanding:} Enhanced co-temporal reasoning improves the model's ability to comprehend concurrent events, leading to more accurate identification of event relationships and temporal sequences.
\item \textbf{Timeline Generation:} Improved co-temporal reasoning facilitates more precise timeline generation by better understanding the relationships between events, resulting in more coherent and accurate timelines.
\end{itemize}

\section{Conclusion} 
In this paper, we propose the \cotempqa datasets to facilitate the investigation of
under-explored co-temporal reasoning problems for large language models. Extensive experiments have shown a significant gap between existing advanced LLMs and human-level performance, even with the enhancement of reasoning approaches. We also discover that mathematical reasoning is crucial for understanding co-temporal events and propose a math-based strategy to improve LLMs' co-temporal reasoning.
Reasoning on concurrent and intricate temporal relations remains an open research question, and we hope more enhancement to develop upon our \cotempqa datasets.

\section*{Limitations}

There are still some limitations in our work, which are listed below:
\begin{itemize}[leftmargin=*]
\setlength{\itemsep}{0pt}
    \item While the \cotempqa is comprehensive, with 4,748 samples, a larger dataset could provide more robust evaluation and training opportunities. As our dataset construction pipeline is adaptable to more data sources besides Wikidata, we will continuously expand our dataset and explore model training in future versions.
    \item For our open-book QA setting, we directly provide the subject's relevant facts in a structured format in the prompt. Recent work shows that LLM's performance in context-based reasoning was significantly weaker than in the former~\cite{chu2023timebench}. In the future, we will employ some retrieval tools to construct prompts with more contextually rich information sources.
    \item We evaluate the co-temporal reasoning capabilities from the perspective of task performance. However, a more direct approach could involve analyzing how the model's neurons and hidden states are triggered~\cite{zhang2023multimodal}. This limitation is not unique to our study and is common in most evaluations of Large Language Models.
\end{itemize}

\section*{Acknowledgement}
We want to thank all the anonymous reviewers for their valuable comments. This work was supported by the National Science Foundation of China (NSFC No. 62206194), the Priority Academic Program Development of Jiangsu Higher Education Institutions, the Natural Science Foundation of Jiangsu Province, China (Grant No. BK20220488), Young Elite Scientists Sponsorship Program by CAST (2023QNRC001), and Supercomputing Center in Yancheng, Grant No.
20231001.

\bibliography{anthology,custom}
\clearpage
\appendix

\section{Implementation Details}
\label{apx:implementatio_details}
We utilize the OpenAI API\footnote{https://platform.openai.com/} to evaluate all closed-source models, and for open-source models, we employ the transformers library~\cite{wolf-etal-2020-transformers}. In all our experiments, we set the temperature to 0 and the maximum length to 256.
These experiments were conducted across a full range of scales for each evaluated model.

\section{The Details of Co-temporal Extraction}
\label{apx:extraction_details}
Building on our approach to structuring time-dependent facts from Wikidata, we delve into extracting co-temporal facts by identifying overlaps in the timestamps of different facts. Specifically, we compare a given fact $f_{i} = (s_i, r_i, o_i, t_{s_{i}}, t_{e_{i}})$ with another fact $f_{j}$ in five distinct scenarios:

\subsection{Scenario 1: \texorpdfstring{$(\mathcal{S}, \mathcal{R}, \mathcal{\overline{O}})$}{si, ri, oj}}
\textbf{Original Fact:} $f_{i}$ as defined above. \\
\textbf{Compared Fact:} $f_{j} = (s_i, r_i, o_j, t_{s_{j}}, t_{e_{j}})$. \\
\textbf{Explanation:} In this scenario, the subject $s_i$ and relation $r_i$ remain constant, indicating the same subject in the same type of relationship. However, the object changes, where the subject is related to different objects during co-temporal periods.\\
\textbf{Template:} While <$subject1$> was holding the position of <$object1$>, which position did <$subject1$> hold during the same time span?\\

\subsection{Scenario 2: \texorpdfstring{$(\mathcal{S}, \mathcal{\overline{R}}, \mathcal{\overline{O}})$}{si, rj, oj}}
\textbf{Original Fact:} $f_{i}$ as above. \\
\textbf{Compared Fact:} $f_{j} = (s_i, r_j, o_j, t_{s_{j}}, t_{e_{j}})$. \\
\textbf{Explanation:} Here, the subject $s_i$ stays constant while the relation and the object change. This scenario is crucial for identifying instances where a single subject is involved in different relationships with different objects concurrently.\\
\textbf{Template:} While <$subject1$> was holding the position of <$object1$>, which political party did <$subject1$> belong to simultaneously? \\

\subsection{Scenario 3: \texorpdfstring{$(\mathcal{\overline{S}}, \mathcal{R}, \mathcal{O})$}{sj, ri, oi}}
\textbf{Original Fact:} $f_{i}$ as above. \\
\textbf{Compared Fact:} $f_{j} = (s_j, r_i, o_i, t_{s_{j}}, t_{e_{j}})$. \\
\textbf{Explanation:} This scenario reflects cases where the relationship and object remain constant, but the subject changes. It suggests different subjects simultaneously having the same type of relationship with the same object. \\
\textbf{Template:} While <$subject1$> was holding the position of <$object1$>, who also held the position of <$object1$> concurrently?\\

\subsection{Scenario 4: \texorpdfstring{$(\mathcal{\overline{S}}, \mathcal{R}, \mathcal{\overline{O}})$}{sj, ri, oj}} 
\textbf{Original Fact:} $f_{i}$ as above. \\
\textbf{Compared Fact:} $f_{j} = (s_j, r_i, o_j, t_{s_{j}}, t_{e_{j}})$. \\
\textbf{Explanation:} Only the relationship remains constant in this case, while both the subject and the object change. This scenario signifies instances where different subjects share a common relationship with different objects concurrently. \\
\textbf{Template:} While <$subject1$> was playing for <$object1$>, which team did <$subject2$> play for within the same time interval? \\

\subsection{Scenario 5: \texorpdfstring{$(\mathcal{\overline{S}}, \mathcal{\overline{R}}, \mathcal{\overline{O}})$}{sj, rj, oj}} 
\textbf{Original Fact:} $f_{i}$ as above. \\
\textbf{Compared Fact:} $f_{j} = (s_j, r_j, o_j, t_{s_{j}}, t_{e_{j}})$. \\
\textbf{Explanation:} This scenario represents completely distinct facts that overlap in time, with all quintuplet elements changing.\\
\textbf{Template:} While <$subject1$> was holding the position of <$object1$>, which employer did <$subject2$> work for during the same time?\\

\section{Case Study}
\label{apx:case_study}
\begin{table*}[t!]
\resizebox{\linewidth}{!}{
    \centering
\begin{tabular}{l|l}
\toprule
\textbf{Equal}               & 
\begin{tabular}[c]{@{}l@{}}\textbf{Context:} \\ Thomas Wenski holds the position of auxiliary bishop \textbf{in June 24, 1997}. \\ Thomas Wenski holds the position of titular bishop \textbf{in June 24, 1997}. \\   ......  \\ \textbf{Question:} \\ While Thomas Wenski was holding the position of auxiliary bishop, \\ which position did Thomas Wenski during the same time period?
\end{tabular}                                                                                                                                                                                                                                                                                                                                                                              \\ \hline
\textbf{Overlap}                & 
\begin{tabular}[c]{@{}l@{}}\textbf{Context:} \\ Avet Ter-Gabrielyan works for Yerevan Komitas State Conservatory \textbf{from 1923 to 1944}. \\ Avet Ter-Gabrielyan works for Komitas Quartet \textbf{from 1924 to 1976}. \\  ......  \\ \textbf{Question:} \\ While Avet Ter-Gabrielyan was working for Yerevan Komitas State Conservatory, \\which employer did Avet Ter-Gabrielyan work for during the same time span?
\end{tabular}           \\ \hline
\textbf{During}            & \begin{tabular}[c]{@{}l@{}}\textbf{Context:} \\ Yaiza Canzani works for Institute for Advanced Study \textbf{from July 1, 2014 to July 31, 2015}. \\ Yaiza Canzani works for Harvard University \textbf{from June 30, 2013 to June 30, 2016}. \\  ......  \\ \textbf{Question:} \\ While Yaiza Canzani was working for Institute for Advanced Study, \\ which employer did Yaiza Canzani work for at the same time?
\end{tabular} \\ \hline

\textbf{Mix}            & \begin{tabular}[c]{@{}l@{}}\textbf{Context:} \\ John Daniel FitzGerald holds the position of Minister for Justice from \textbf{July 23, 1919 to April 12, 1920}. \\ John Daniel FitzGerald holds the position of Minister for Local Government \textbf{from November 15, 1916 to April 12, 1920}. \\ John Daniel FitzGerald holds the position of Solicitor General for New South Wales \textbf{from July 23, 1919 to April 12, 1920}. \\ John Daniel FitzGerald holds the position of Vice-President of the Executive Council \textbf{from April 27, 1915 to July 30, 1919}. \\ ......  \\ \textbf{Question:} \\ While John Daniel FitzGerald was holding the position of Minister for Justice,  \\ which position did John Daniel FitzGerald during the same time period?
\end{tabular} \\ 
\bottomrule
\end{tabular}}
    \caption{Case Study. We provide some representative examples to give an intuitive presentation of the varying difficulties in the \cotempqa. Time periods are highlighted in \textbf{bold} for easy identification.}
    \label{tb:case-study}
\end{table*}

Table~\ref{table:main_table} indicates that existing LLMs can effectively reason about straightforward concurrent events. However, they encounter difficulties in more complex tasks that require a deeper understanding and comprehension of co-temporal reasoning.
In this section, we provide a further case study to show this difference.
As shown in Table~\ref{tb:case-study}, the \textbf{Equal} scenario is more accessible for LLMs as their co-temporal time interval entirely overlap. \textbf{Overlap} and \textbf{During} scenarios present intricate temporal intersections, necessitating more implicit reasoning to understand the co-temporal relationships.
It becomes more challenging to determine whether one time period intersects another (i.e., During and Overlap) compared to the straightforward identification in the \textbf{Equal} scenario.
Additionally, the \textbf{Mix} scenario has several correct answers and contains various co-temporal relationships, which makes it the most challenging compared to other scenarios.

\section{Further Analysis for the Mix Scenario}
\label{apx:mix_analysis}
\begin{table*}[t!]
\resizebox{\linewidth}{!}{
\begin{tabular}{lccc|c}
\toprule
\textbf{Model}                    & \textbf{Precision}    & \textbf{Recall}     & \textbf{F1} & \textbf{Prediction} \\
\midrule
\textbf{CodeLLaMA-34B}            & 41.6    & 47.3   & 40.3  & Minister of Finance, Minister of Education of New Zealand, Minister of Justice   \\
\textbf{LLaMA-70B}                & 41.3    & \textbf{58.3}      & \textbf{42.5} & Minister of Finance, Minister of Education of New Zealand, Minister of Justice     \\
\textbf{WizardMath-70B}           & \textbf{47.5}    & 28.8      & 32.9  & Minister of Education of New Zealand \\
\bottomrule
\end{tabular}}
\caption{The performance of different open source model in the mixed scenario and the models' prediction when the ground truth is \textit{Minister of Finance, Minister of Education of New Zealand.}}
\label{tab:mix_analysis}
\end{table*}
In this section, we provide further analysis for WizardMath's reduced effectiveness in the \textbf{Mix} scenario by the case and experimental results. As shown in Table~\ref{tab:mix_analysis}, CodeLLaMA and LLaMA prefer to provide all potential answers, but WizardMath only returns a signal alternative answer.
WizardMath is trained to return the answer with the highest probability since preciseness and accuracy are required in mathematical reasoning~\cite{lu-etal-2023-survey}. As questions have more than one correct answers in the \textbf{Mix} scenario, WizardMath score the highest precision~(47.5\%) and the lowest Recall~(28.8\%), leading to underperformance in this scenario.

\section{Error Analysis}
\label{apx:error_analysis}
\begin{table*}[t!]
\resizebox{\linewidth}{!}{
    \centering
\begin{tabular}{l|l}
\toprule
\textbf{Error Type} & \textbf{Example} \\ \midrule
\textbf{Incomplete answer errors~(27.93\%)}               & 
\begin{tabular}[c]{@{}l@{}}\textbf{Question:} \\ While Bodil Nyboe Andersen was working for Tryg, \\ which employer did Bodil Nyboe Andersen work for within the same time interval? \\ \textbf{Gold answer:}\\ Sampension, Alka \\  \textbf{Predict answer:} \\ Alka
\end{tabular}           \\ \hline

\textbf{Uncertainty errors~(43.14\%)}               & 
\begin{tabular}[c]{@{}l@{}}\textbf{Question:} \\ While Alain Decaux was holding the position of director, \\ which position did Alain Decaux at the same time? \\ \textbf{Gold answer:} \\ president \\ \textbf{Predict answer:} \\ Alain Decaux did not hold any other position at the same time he was a director from 1969 to 1971.
\end{tabular}           \\ \hline

\textbf{Incorrect answer errors~(28.93\%)}               & 
\begin{tabular}[c]{@{}l@{}}\textbf{Question:} \\ While Thomas Wenski was holding the position of auxiliary bishop, \\ which position did Thomas Wenski during the same time period? \\ \textbf{Gold answer:} \\ titular bishop \\ \textbf{Predict answer:} \\ Minister for Children (Denmark)
\end{tabular}           \\
\bottomrule
\end{tabular}}
    \caption{Case-wise error analysis. The incorrect answers are categorized into three types~(i.e., incomplete answer errors, uncertainty errors, incorrect answer errors). For each type, an illustrative example is provided to enhance clarity and understanding. ``Uncertainty errors'' are the most frequent error type.}
    \label{tb:error_analysis}
\end{table*}

To better understand the mistakes made by models, we focused our investigation on the responses generated by GPT-4 under 0-shot CoT.
We divide the errors into three categories:
\begin{itemize}[leftmargin=*]
\setlength{\itemsep}{0pt}
    \item \textbf{Incomplete answer errors} refer to situations where the questions have multiple correct answers, but failing to return all of them.
    \item \textbf{Uncertainty errors} represent the models' inability to extract the co-temporal relation from the context provided and refuse to response the question.
    \item \textbf{Incorrect answer errors} are characterized by the model cannot return the correct answers, which means the models are insufficient in co-temporal reasoning.
\end{itemize}
Our case-wise error analysis is shown in Table~\ref{tb:error_analysis}, ``uncertainty errors'' are the most frequent error type, accounting for 43.14\%.
We assume that the GPT-4 tends to provide relatively conservative responses and only returns answers when there is a certain level of confidence~\cite{cheng2024ai}.
Future research needs to optimize the model's framework and further enhance the capabilities of LLMs in co-temporal understanding and reasoning.

\section{Prompts}
\label{apx:prompt}
The prompts and demonstrations can be found from Figure~\ref{fig:start} to Figure~\ref{fig:end}.

\begin{table*}[t]
\begin{small}
    \centering
    \small
    \renewcommand{\arraystretch}{1.25}
    \resizebox{\linewidth}{!}{
    \begin{tabular}{lccccccccccccccc|c}
         \toprule
        \multirow{2}{*}{\bf Model} &
        \multicolumn{3}{c}{$(\mathcal{S}, \mathcal{R}, \mathcal{\overline{O}})$} & \multicolumn{3}{c}{$(\mathcal{S}, \mathcal{\overline{R}}, \mathcal{\overline{O}})$} & \multicolumn{3}{c}{$(\mathcal{\overline{S}}, \mathcal{R}, \mathcal{O})$} & \multicolumn{3}{c}{$(\mathcal{\overline{S}}, \mathcal{R}, \mathcal{\overline{O}})$} & \multicolumn{3}{c|}{$(\mathcal{\overline{S}}, \mathcal{\overline{R}}, \mathcal{\overline{O}})$} & \multirow{2}{*}{\textbf{Overall}}\\
        \cmidrule(lr){2-4} \cmidrule(lr){5-7} \cmidrule(lr){8-10} \cmidrule(lr){11-13} \cmidrule(lr){14-16}
        & \bf Acc. & \bf $\textrm{F}_1$ & \bf Avg. & \bf Acc. & \bf $\textrm{F}_1$ & \bf Avg. & \bf Acc. & \bf $\textrm{F}_1$ & \bf Avg. & \bf Acc. &  \bf $\textrm{F}_1$ & \bf Avg. & \bf Acc. & \bf $\textrm{F}_1$ & \bf Avg. \\
        \hline
        \multicolumn{17}{c}{\bf{{The Closed Book Question Answer~(CBQA) setting}}} \\ 
        \textsc{GPT-3.5-Turbo-0613} & \bf 7.3 & \bf 11.9 & \bf 9.6 & 29.7 & 31.1 & 30.4 & \bf 3.0 & \bf 4.5 & \bf 3.7 & 13.3 & \bf 30.6 & \bf 22.0 & 8.5 & \bf 21.8 & \bf 15.1 & \bf 16.3\\
        \textsc{GPT-4-0613} & 3.5 & 8.0 & 5.7 & \bf 30.0 & \bf 31.3 & \bf 30.7 & 2.3 & 4.4 & 3.3 & \bf 15.0  & 24.5 & 19.7 & \bf 9.5 & 15.1 & 12.3 & 14.5\\     
        \specialrule{0.04em}{1pt}{1pt}
        \multicolumn{17}{c}{\bf{{The Open Book Question Answer~(OBQA) setting}}} \\ 
        \textsc{GPT-3.5-Turbo} & 34.2 & 55.8 & 45.0 & 54.7 & 57.1 & 55.9 & 34.4 & 55.6 & 45.0 & 17.8 & 33.5 & 25.6 & 15.1 & 28.0 & 21.5 & 38.9\\
        \textsc{GPT-4} & \textbf{57.0} & \textbf{74.7} & \textbf{65.9} & \textbf{68.0} & \textbf{70.8} & \textbf{69.4} & \textbf{58.0} & \textbf{73.3} & \textbf{65.6} & \textbf{29.9} &
        \textbf{48.8} & \textbf{39.4} &
        \textbf{23.5} & \textbf{39.8} &
        \textbf{31.7} & \textbf{54.7} \\ 
        \specialrule{0.04em}{1pt}{1pt}
        \textsc{CodeLLaMa-7b} & 4.3 & \bf 24.3 & \bf 14.3 & 6.5 & 23.6 & 15.0 & 1.3 & 15.4 & 8.4  & 1.7 & 9.9 & 5.8  & 2.2 & 14.3 & 8.3  & 10.5   \\
        \textsc{WizardCoder-7b}  & 3.3 & 17.9 & 10.6 & 16.6 & \bf 27.3 & 22.0 & 3.6 & 16.7 & 10.1 & 1.7 & 8.2 & 4.9 & 2.3 & 12.5 & 7.4 & 11.2   \\
        \textsc{LLaMa-7b} & 2.5 & 20.5 & 11.5 & 9.5 & 23.0 & 16.3 & 1.9 & \bf 20.5 & 11.2 & 2.2 & 16.8 & 9.5 & 3.4 & \bf 18.7 & 11.1 & 12.0   \\
        \textsc{WizardMath-7b} & \bf 7.2 & 16.8 & 12.0 & \bf 22.8 & 26.6 & \bf 24.7 & \bf 4.5 & 19.1 & \bf 11.8 & \bf 7.5 & \bf 17.9 & \bf 12.7 & \bf 7.1 & 17.1 & \bf 12.1 & \bf 14.8   \\
        \specialrule{0.04em}{1pt}{1pt}
        \textsc{CodeLLaMa-13b} & 6.0 & 26.0 & 16.0 & 7.7 & 26.9 & 17.3 & 2.0 & 17.9 & 10.0 & 1.0 & 16.2 & 8.6 & 1.7 & 16.7 & 9.2 & 12.4   \\
        \textsc{WizardCoder-13b} & 5.0 & 16.6 & 10.8 & 19.6 & 30.2 & 24.9 & \bf 3.9 & \bf 23.3 & \bf 13.6 & \bf 6.3 & 15.1 & \bf 10.7 & \bf 4.6 & 12.6 & 8.6 & 13.9   \\
        \textsc{LLaMa-13b} & 6.2 & \bf 28.6 & \bf 17.4 & 10.5 & 27.0 & 18.7 & 3.6 & 20.9 & 12.2 & 2.3 & \bf 17.2 & 9.8 & 3.0 & \bf 17.6 & \bf 10.3 & 13.8   \\
        \textsc{WizardMath-13b} & \bf 11.2 & 19.7 & 15.5 & \bf 30.9 & \bf 34.0 & \bf 32.5 & 3.1 & 9.1 & 6.1 & 5.6 & 13.8 & 9.7 & 4.0 & 9.1 & 6.5 & \bf 14.4   \\
        \specialrule{0.04em}{1pt}{1pt}
        \textsc{CodeLLaMa-34b} & 9.9 & 42.4 & 26.2 & 19.6 & 38.3 & 29.0 & 4.7 & 27.1 & 15.9 & 4.3 & 25.6 & 14.9 & 3.6 & 21.6 & 12.6 & 20.0   \\
        \textsc{WizardCoder-34b} & 11.0 & 22.6 & 16.8 & 35.2 & 38.0 & 36.6 & \bf 9.0 & 27.4 & 18.2 & 7.5 & 15.9 & 11.7 & 7.4 & 16.0 & 11.7 & 19.2   \\
        \textsc{LLaMa-70b} & 10.3 & \bf 43.6 & 26.9 & 14.7 & 37.6 & 26.1 & 7.1 & \bf 31.7 & \bf 19.4 & 7.0 & 32.4 & 19.7 & 6.2 & 29.9 & 18.1 & 22.2   \\
        \textsc{WizardMath-70b} & \bf 18.3 & 37.8 & \bf 28.1 & \bf 49.8 & \bf 53.4 & \bf 51.6 & 8.6 & 24.8 & 16.7 & \bf 20.1 & \bf37.3 & \bf 28.7 & \bf 17.4 & \bf 30.1 & \bf 23.8 & \bf 30.1   \\

        \bottomrule
        \end{tabular}}
    \caption{Experimental results of different triple element types in \cotempqa. The best performance is \textbf{bold}.} 
    \label{table:extensive_SRO}
\end{small}
\end{table*}

\begin{sidewaystable}[t]
\centering
\resizebox{\linewidth}{!}{
\begin{tabular}{@{}clcl@{}}
\toprule
\textbf{WikiData ID} & \textbf{KB Relation Pairs}              & \textbf{\# Queries} & \textbf{Template}   
 \\ \midrule

P102-P102   & political party \& political party                & 475      & While \textless{}subject\textgreater~  was a member of <object>, which political party did \textless{}subject\textgreater~  belong to within the same time interval? \\
P39-P39     & position held \& position held                    & 1,017    & While  \textless{}subject\textgreater~ was holding the position of <object>, which position did \textless{}subject\textgreater~ hold during the same time span? \\
P108-P108   & employer \& employer                              & 768      & While \textless{}subject\textgreater~ was working for <object>, which employer did \textless{}subject\textgreater~ work for during the same time period? \\
P54-P54     & member of sports team \& member of sports team    & 204      & While \textless{}subject\textgreater~ was playing for <object>, which team did \textless{}subject\textgreater~ play for at the same time? \\
P69-P69     & educated at \& educated at                        & 258      & While  \textless{}subject\textgreater~ attended <object>, which school was  \textless{}subject\textgreater~ attending during the identical time period? \\
P127-P127   & owned by \& owned by                              & 75       & While \textless{}subject\textgreater~ was owned by <object>, who was the owner of \textless{}subject\textgreater~ concurrently? \\
P102-P39    & political party \& position held                  & 117      & While \textless{}subject\textgreater~  was a member of <object>, which position did \textless{}subject\textgreater~ hold simultaneously? \\
P102-P108   & political party \& employer                       & 101      & While \textless{}subject\textgreater~  was a member of <object>, which employer did \textless{}subject\textgreater~ work for during the same time span? \\
P102-P69    & political party \& educated at                    & 74       & While \textless{}subject\textgreater~  was a member of <object>, which school was  \textless{}subject\textgreater~ attending within the same time interval? \\
P39-P102    & position held \& political party                  & 420      & While  \textless{}subject\textgreater~ was holding the position of <object>, which political party did \textless{}subject\textgreater~  belong to during the same time period? \\
P39-P108    & position held \& employer                         & 380      & While  \textless{}subject\textgreater~ was holding the position of <object>, which employer did \textless{}subject\textgreater~ work for at the same time? \\
P108-P39    & employer \& position held                         & 125      & While \textless{}subject\textgreater~ was working for <object>, which position did \textless{}subject\textgreater~ hold during the identical time period? \\
P108-P69    & employer \& educated at                           & 241      & While \textless{}subject\textgreater~ was working for <object>, which school was  \textless{}subject\textgreater~ attending concurrently? \\
P54-P69     & member of sports team \& educated at              & 77       & While \textless{}subject\textgreater~ was playing for <object>, which school was  \textless{}subject\textgreater~ attending simultaneously? \\
P69-P102    & educated at \& political party                    & 187      & While  \textless{}subject\textgreater~ attended <object>, which political party did \textless{}subject\textgreater~  belong to during the same time span? \\
P69-P39     & educated at \& position held                      & 95       & While  \textless{}subject\textgreater~ attended <object>, which position did \textless{}subject\textgreater~ hold within the same time interval? \\
P69-P108    & educated at \& employer                           & 134      & While  \textless{}subject\textgreater~ attended <object>, which employer did \textless{}subject\textgreater~ work for during the same time period? \\ \bottomrule
\end{tabular}}
\caption{Templates used for converting Wikidata facts into natural questions. We manually predefine 17 types of KB relation pairs and ensure these relations are interconnected logically and contextually. Taking the ``position held \& employer'' relation pair as an example, understanding the overlap between the period when a person held a specific position and their employment at the same organization can provide valuable insights into career patterns.}
\label{tab:question-templates}
\end{sidewaystable}
\clearpage

\begin{figure*}[ht]
    \centering
    \footnotesize
    \begin{tabularx}{\linewidth}{|X|}
    \toprule
        \textbf{Question}: While Valdis Dombrovskis was holding the position of European Commissioner for Trade, which position did Valdis Dombrovskis during the identical time period?\\
        \textbf{Only return the answer}:  \\
    \bottomrule
    \end{tabularx}
    \caption{Default prompt for \textbf{C}losed-\textbf{B}ook \textbf{QA}~(\textbf{CBQA}) in our proposed \cotempqa ~}
    \label{fig:start}
\end{figure*}

\begin{figure*}[h]
    \centering
    \footnotesize
    \begin{tabularx}{\linewidth}{|X|}
    \toprule
        \textbf{Question}: While Valdis Dombrovskis was holding the position of European Commissioner for Trade, which position did Valdis Dombrovskis during the identical time period?\\
        \textbf{Answer: Let's think step by step,}  \\
    \bottomrule
    \end{tabularx}
    \caption{Zero-cot prompt for \textbf{C}losed-\textbf{B}ook \textbf{QA}~(\textbf{CBQA}) in our proposed \cotempqa ~}
    \label{fig:prompt-demos-timeqa}
\end{figure*}

\begin{figure*}[h]
    \centering
    \footnotesize
    \begin{tabularx}{\linewidth}{|X|}
    \toprule
        \textbf{Question}: While Valdis Dombrovskis was holding the position of European Commissioner for Trade, which position did Valdis Dombrovskis during the identical time period?\\
        \textbf{Only return the answer}: European Commissioner for Internal Market and Services \\
        ......\\
        \textbf{Question}: While Eduard Jan Dijksterhuis was working for Leiden University, which employer did Eduard Jan Dijksterhuis work for during the same time span?\\
        \textbf{Only return the answer}: \\
        
    \bottomrule
    \end{tabularx}
    \caption{Few-shot prompt for \textbf{C}losed-\textbf{B}ook \textbf{QA}~(\textbf{CBQA}) in our proposed \cotempqa ~(5-shot)}
    \label{fig:prompt-demos-timeqa}
\end{figure*}

\begin{figure*}[h]
    \centering
    \footnotesize
    \begin{tabularx}{\linewidth}{|X|}
    \toprule
        \textbf{Question}: While Valdis Dombrovskis was holding the position of European Commissioner for Trade, which position did Valdis Dombrovskis during the identical time period?\\
        \textbf{Answer}: According to the fact, Valdis Dombrovskis became the European Commissioner for Trade on August 26, 2020. He also held the position of European Commissioner for Internal Market and Services from July 16, 2016, to October 12, 2020. This period overlaps with his tenure as Commissioner for Trade. Therefore, the answer is European Commissioner for Internal Market and Services. \\
        ......\\
        \textbf{Question}: While Eduard Jan Dijksterhuis was working for Leiden University, which employer did Eduard Jan Dijksterhuis work for during the same time span?\\
        \textbf{Answer: According to the fact,} \\
        
    \bottomrule
    \end{tabularx}
    \caption{Few-shot\&CoT prompt for \textbf{C}losed-\textbf{B}ook \textbf{QA}~(\textbf{CBQA}) in our proposed \cotempqa ~(5-shot)}
\end{figure*}

\begin{figure*}[h]
    \centering
    \footnotesize
    \begin{tabularx}{\linewidth}{|X|}
    \toprule
        \textbf{Question}: While Valdis Dombrovskis was holding the position of European Commissioner for Trade, which position did Valdis Dombrovskis during the identical time period?\\
        \textbf{Answer}: According to the context, Valdis Dombrovskis became the European Commissioner for Trade on August 26, 2020. The datetime can be formed (2020, 8, 26).\\ The content provided and related to the question can be structured as:\\(Vice-President of the European Commission, (2019, 12, 1)).\\(European Commissioner for Internal Market and Services, (2016, 6, 16), (2020, 10, 12)).\\(European Commissioner for An Economy, (2019, 10, 1)).\\(Prime Minister of Latvia, (2009, 3, 12), (2014, 1, 22)).\\(Minister of Finance, (2002, 11, 7), (2004, 3, 9)).\\Given the (2020, 8, 26), compared with all contents related, we find that $[(2016, 6, 16)-(2020, 10, 12)] \cap (2020, 8, 26) \neq \emptyset$.\\Therefore the answer is European Commissioner for Internal Market and Services.\\
        ......\\
        \textbf{Question}: While Eduard Jan Dijksterhuis was working for Leiden University, which employer did Eduard Jan Dijksterhuis work for during the same time span?\\
        \textbf{Answer: According to the fact,} \\
        
    \bottomrule
    \end{tabularx}
    \caption{Few-shot\&Mr-CoT prompt for \textbf{C}losed-\textbf{B}ook \textbf{QA}~(\textbf{CBQA}) in our proposed \cotempqa ~(5-shot)}
\end{figure*}

\begin{figure*}[ht]
    \centering
    \footnotesize
    \begin{tabularx}{\linewidth}{|X|}
    \toprule
        \textbf{Answer the question based on the context:}\\
        \textbf{Context}: Valdis Dombrovskis holds the position of Vice-President of the European Commission in December 1, 2019.\\ Valdis Dombrovskis holds the position of European Commissioner for Internal Market and Services from July 16, 2016 to October 12, 2020.\\ Valdis Dombrovskis holds the position of European Commissioner for Trade in August 26, 2020.\\ Valdis Dombrovskis holds the position of European Commissioner for An Economy that Works for People in December 1, 2019.\\ Valdis Dombrovskis holds the position of Prime Minister of Latvia from March 12, 2009 to January 22, 2014.\\ Valdis Dombrovskis holds the position of Minister of Finance from November 7, 2002 to March 9, 2004.\\
        \textbf{Question}: While Valdis Dombrovskis was holding the position of European Commissioner for Trade, which position did Valdis Dombrovskis during the identical time period?\\
        \textbf{Only return the answer}:  \\
    \bottomrule
    \end{tabularx}
    \caption{Default prompt for \textbf{O}pen-\textbf{B}ook \textbf{QA}~(\textbf{OBQA}) in our proposed \cotempqa ~}
    \label{fuck}
\end{figure*}

\begin{figure*}[h]
    \centering
    \footnotesize
    \begin{tabularx}{\linewidth}{|X|}
    \toprule
        \textbf{Answer the question based on the context:}\\
        \textbf{Context}: Valdis Dombrovskis holds the position of Vice-President of the European Commission in December 1, 2019.\\ Valdis Dombrovskis holds the position of European Commissioner for Internal Market and Services from July 16, 2016 to October 12, 2020.\\ Valdis Dombrovskis holds the position of European Commissioner for Trade in August 26, 2020.\\ Valdis Dombrovskis holds the position of European Commissioner for An Economy that Works for People in December 1, 2019.\\ Valdis Dombrovskis holds the position of Prime Minister of Latvia from March 12, 2009 to January 22, 2014.\\ Valdis Dombrovskis holds the position of Minister of Finance from November 7, 2002 to March 9, 2004.\\
        \textbf{Question}: While Valdis Dombrovskis was holding the position of European Commissioner for Trade, which position did Valdis Dombrovskis during the identical time period?\\
        \textbf{Answer: Let's think step by step,}  \\
    \bottomrule
    \end{tabularx}
    \caption{Zero-cot prompt for \textbf{O}pen-\textbf{B}ook \textbf{QA}~(\textbf{OBQA}) in our proposed \cotempqa ~}
    \label{fig:prompt-demos-timeqa}
\end{figure*}

\begin{figure*}[h]
    \centering
    \footnotesize
    \begin{tabularx}{\linewidth}{|X|}
    \toprule
        \textbf{Answer the question based on the context:}\\
        \textbf{Context}: Valdis Dombrovskis holds the position of Vice-President of the European Commission in December 1, 2019.\\ Valdis Dombrovskis holds the position of European Commissioner for Internal Market and Services from July 16, 2016 to October 12, 2020.\\ Valdis Dombrovskis holds the position of European Commissioner for Trade in August 26, 2020.\\ Valdis Dombrovskis holds the position of European Commissioner for An Economy that Works for People in December 1, 2019.\\ Valdis Dombrovskis holds the position of Prime Minister of Latvia from March 12, 2009 to January 22, 2014.\\ Valdis Dombrovskis holds the position of Minister of Finance from November 7, 2002 to March 9, 2004.\\
        \textbf{Question}: While Valdis Dombrovskis was holding the position of European Commissioner for Trade, which position did Valdis Dombrovskis during the identical time period?\\
        \textbf{Only return the answer}: European Commissioner for Internal Market and Services \\
        ......\\
        \textbf{Answer the question based on the context:}\\
        \textbf{Context}: Eduard Jan Dijksterhuis attended University of Groningen from 1911 to 1918.\\Eduard Jan Dijksterhuis worked for Duke University School of Medicine from August 28, 1912 to April 28, 1923.\\Eduard Jan Dijksterhuis works for Leiden University from July 5, 1954 to September 5, 1960.\\Eduard Jan Dijksterhuis worked for American University of Armenia in August, 1911.\\Eduard Jan Dijksterhuis worked for Austin College from July, 1936 to April, 1947.\\Eduard Jan Dijksterhuis worked for Sonoma State University in July, 1932.\\Eduard Jan Dijksterhuis worked for Fairfax Media in December 16, 1942.\\Eduard Jan Dijksterhuis worked for Canadian Broadcasting Corporation in 1941.\\Eduard Jan Dijksterhuis works for Utrecht University from May 1, 1953 to September 1, 1960.\\Eduard Jan Dijksterhuis worked for Jean-Marie Le Pen in January, 1931.\\
        \textbf{Question}: While Eduard Jan Dijksterhuis was working for Leiden University, which employer did Eduard Jan Dijksterhuis work for during the same time span?\\
        \textbf{Only return the answer}: \\
        
    \bottomrule
    \end{tabularx}
    \caption{Few-shot prompt for \textbf{O}pen-\textbf{B}ook \textbf{QA}~(\textbf{OBQA}) in our proposed \cotempqa ~(5-shot)}
    \label{fig:prompt-demos-timeqa}
\end{figure*}

\begin{figure*}[h]
    \centering
    \footnotesize
    \begin{tabularx}{\linewidth}{|X|}
    \toprule
        \textbf{Answer the question based on the context:}\\
        \textbf{Context}: Valdis Dombrovskis holds the position of Vice-President of the European Commission in December 1, 2019.\\ Valdis Dombrovskis holds the position of European Commissioner for Internal Market and Services from July 16, 2016 to October 12, 2020.\\ Valdis Dombrovskis holds the position of European Commissioner for Trade in August 26, 2020.\\ Valdis Dombrovskis holds the position of European Commissioner for An Economy that Works for People in December 1, 2019.\\ Valdis Dombrovskis holds the position of Prime Minister of Latvia from March 12, 2009 to January 22, 2014.\\ Valdis Dombrovskis holds the position of Minister of Finance from November 7, 2002 to March 9, 2004.\\
        \textbf{Question}: While Valdis Dombrovskis was holding the position of European Commissioner for Trade, which position did Valdis Dombrovskis during the identical time period?\\
        \textbf{Answer}: According to the context, Valdis Dombrovskis became the European Commissioner for Trade on August 26, 2020. He also held the position of European Commissioner for Internal Market and Services from July 16, 2016, to October 12, 2020. This period overlaps with his tenure as Commissioner for Trade. Therefore, the answer is European Commissioner for Internal Market and Services. \\
        ......\\
        \textbf{Answer the question based on the context:}\\
        \textbf{Context}: Eduard Jan Dijksterhuis attended University of Groningen from 1911 to 1918.\\Eduard Jan Dijksterhuis worked for Duke University School of Medicine from August 28, 1912 to April 28, 1923.\\Eduard Jan Dijksterhuis works for Leiden University from July 5, 1954 to September 5, 1960.\\Eduard Jan Dijksterhuis worked for American University of Armenia in August, 1911.\\Eduard Jan Dijksterhuis worked for Austin College from July, 1936 to April, 1947.\\Eduard Jan Dijksterhuis worked for Sonoma State University in July, 1932.\\Eduard Jan Dijksterhuis worked for Fairfax Media in December 16, 1942.\\Eduard Jan Dijksterhuis worked for Canadian Broadcasting Corporation in 1941.\\Eduard Jan Dijksterhuis works for Utrecht University from May 1, 1953 to September 1, 1960.\\Eduard Jan Dijksterhuis worked for Jean-Marie Le Pen in January, 1931.\\
        \textbf{Question}: While Eduard Jan Dijksterhuis was working for Leiden University, which employer did Eduard Jan Dijksterhuis work for during the same time span?\\
        \textbf{Answer: According to the context,} \\
        
    \bottomrule
    \end{tabularx}
    \caption{Few-shot\&CoT prompt for \textbf{O}pen-\textbf{B}ook \textbf{QA}~(\textbf{OBQA}) in our proposed \cotempqa ~(5-shot)}
\end{figure*}

\begin{figure*}[h]
    \centering
    \footnotesize
    \begin{tabularx}{\linewidth}{|X|}
    \toprule
        \textbf{Answer the question based on the context:}\\
        \textbf{Context}: Valdis Dombrovskis holds the position of Vice-President of the European Commission in December 1, 2019.\\ Valdis Dombrovskis holds the position of European Commissioner for Internal Market and Services from July 16, 2016 to October 12, 2020.\\ Valdis Dombrovskis holds the position of European Commissioner for Trade in August 26, 2020.\\ Valdis Dombrovskis holds the position of European Commissioner for An Economy that Works for People in December 1, 2019.\\ Valdis Dombrovskis holds the position of Prime Minister of Latvia from March 12, 2009 to January 22, 2014.\\ Valdis Dombrovskis holds the position of Minister of Finance from November 7, 2002 to March 9, 2004.\\
        \textbf{Question}: While Valdis Dombrovskis was holding the position of European Commissioner for Trade, which position did Valdis Dombrovskis during the identical time period?\\
        \textbf{Answer}: According to the context, Valdis Dombrovskis became the European Commissioner for Trade on August 26, 2020. The datetime can be formed (2020, 8, 26).\\ The content provided and related to the question can be structured as:\\(Vice-President of the European Commission, (2019, 12, 1)).\\(European Commissioner for Internal Market and Services, (2016, 6, 16), (2020, 10, 12)).\\(European Commissioner for An Economy, (2019, 10, 1)).\\(Prime Minister of Latvia, (2009, 3, 12), (2014, 1, 22)).\\(Minister of Finance, (2002, 11, 7), (2004, 3, 9)).\\Given the (2020, 8, 26), compared with all contents related, we find that $[(2016, 6, 16)-(2020, 10, 12)] \cap (2020, 8, 26) \neq \emptyset$.\\Therefore the answer is European Commissioner for Internal Market and Services.\\
        ......\\
        \textbf{Answer the question based on the context:}\\
        \textbf{Context}: Eduard Jan Dijksterhuis attended University of Groningen from 1911 to 1918.\\Eduard Jan Dijksterhuis worked for Duke University School of Medicine from August 28, 1912 to April 28, 1923.\\Eduard Jan Dijksterhuis works for Leiden University from July 5, 1954 to September 5, 1960.\\Eduard Jan Dijksterhuis worked for American University of Armenia in August, 1911.\\Eduard Jan Dijksterhuis worked for Austin College from July, 1936 to April, 1947.\\Eduard Jan Dijksterhuis worked for Sonoma State University in July, 1932.\\Eduard Jan Dijksterhuis worked for Fairfax Media in December 16, 1942.\\Eduard Jan Dijksterhuis worked for Canadian Broadcasting Corporation in 1941.\\Eduard Jan Dijksterhuis works for Utrecht University from May 1, 1953 to September 1, 1960.\\Eduard Jan Dijksterhuis worked for Jean-Marie Le Pen in January, 1931.\\
        \textbf{Question}: While Eduard Jan Dijksterhuis was working for Leiden University, which employer did Eduard Jan Dijksterhuis work for during the same time span?\\
        \textbf{Answer: According to the context,} \\
        
    \bottomrule
    \end{tabularx}
    \caption{Few-shot\&Mr-CoT prompt for \textbf{O}pen-\textbf{B}ook \textbf{QA}~(\textbf{OBQA}) in our proposed \cotempqa ~(5-shot)}
    \label{fig:end}
\end{figure*}


\end{document}